\documentclass[journal]{IEEEtran}
\usepackage{cite}
\usepackage{amsmath,amssymb,amsfonts}
\usepackage{algorithm}
\usepackage{algpseudocode}
\floatname{algorithm}{Algorithm}
\usepackage{graphicx}
\usepackage{textcomp}
\usepackage{cite}
\usepackage{booktabs}
\usepackage{hyperref}
\usepackage{tabularx}
\usepackage[aboveskip=-5pt,belowskip=0pt]{caption}
\usepackage{multirow}
\usepackage{pifont}
\usepackage{bbm}
\usepackage{bm,upgreek}
\usepackage{verbatim}
\newcommand{\clip}{CLIP}
\newcommand{\mr}[1]{\mathit{#1}}

\newcommand{\loss}{\mathcal{L}}
\newcommand{\mypar}[1]{\vspace{0.35\baselineskip}\noindent\textbf{#1}~}

\usepackage{xcolor}   
\usepackage{colortbl}

\newcommand{\ccol}{\cellcolor{gray!15}}

\newcommand{\fImg}{\mathbf{I}}
\newcommand{\fTxt}{\mathbf{T}}

\newcommand{\xx}{\mathbf{x}}

\newcommand{\real}{\mathbb{R}}

\newcommand{\PP}{\mathbf{P}}
\newcommand{\QQ}{\mathbf{Q}}
\newcommand{\SSS}{\mathbf{S}}

\newcommand{\data}{\mathcal{D}}

\newcommand{\tr}[1]{{#1^{\top}}}

\def\BibTeX{{\rm B\kern-.05em{\sc i\kern-.025em b}\kern-.08em
    T\kern-.1667em\lower.7ex\hbox{E}\kern-.125emX}}

\begin{document}
\title{FAA-CLIP: Federated Adversarial Adaptation of \clip{}}
\author{Yihang~Wu, Ahmad~Chaddad, Christian~Desrosiers, Tareef~Daqqaq,  Reem~Kateb
\thanks{Y. Wu and A. Chaddad are with the School of Artificial Intelligence, Guilin University of Electronic Technology, Guilin, China. \\*Corresponding author: Ahmad Chaddad. 
\\
T. Daqqaq is with the College of Medicine, Taibah University, Al Madinah, Saudi Arabia.\\
R. Kateb is with the College of Computer Science and Engineering, Jeddah University, Jeddah, Saudi Arabia.\\
C. Desrosiers and A. Chaddad are with The Laboratory for Imagery, Vision and Artificial Intelligence, Ecole de Technologie Superieure, Montreal, Canada.
Email: ahmad8chaddad@gmail.com, ahmadchaddad@guet.edu.cn}}

\maketitle

\begin{abstract}
\textcolor{black}{Despite the remarkable performance of vision language models (VLMs) such as Contrastive Language Image Pre-training (CLIP), the large size of these models is a considerable obstacle to their use in federated learning (FL) systems where the parameters of local client models need to be transferred to a global server for aggregation. Another challenge in FL is the heterogeneity of data from different clients, which affects the generalization performance of the solution. In addition, natural pre-trained VLMs exhibit poor generalization ability in the medical datasets, suggests there exists a domain gap. To solve these issues, we introduce a novel method for the Federated Adversarial Adaptation (FAA) of CLIP. Our method, named FAA-CLIP, handles the large communication costs of CLIP using a light-weight feature adaptation module (FAM) for aggregation, effectively adapting this VLM to each client's data while greatly reducing the number of parameters to transfer. By keeping CLIP frozen and only updating the FAM parameters, our method is also computationally efficient. Unlike existing approaches, our FAA-CLIP method directly addresses the problem of domain shifts across clients via a domain adaptation (DA) module. This module employs a domain classifier to predict if a given sample is from the local client or the global server, allowing the model to learn domain-invariant representations. Extensive experiments on six different datasets containing both natural and medical images demonstrate that FAA-CLIP can generalize well on both natural and medical datasets compared to recent FL approaches. Our codes are available at \url{https://github.com/AIPMLab/FAA-CLIP}.}

\end{abstract}

% \textcolor{red}{Finally, decision curve analysis demonstrate that our model is useful for tumor/cancer classification in the 
% clinical environment.}


\section{Introduction}
While models based on deep learning (DL) have achieved ground-breaking results in a broad range of computer vision and natural language understanding tasks, their performance is often dependent on the availability of large datasets \cite{9868083}. In recent years, there has been a growing concern on ensuring data privacy and security, with many organizations implementing regulations and laws such as the EU General Data Protection Regulation (GDPR) \cite{voigt2017eu}. These restrictions on sharing raw data from different organizations poses a siginificant challenge for training robust DL models in fields like medical imaging where privacy is of utmost importance. One of the most promising solutions to this problem is federated learning (FL). FL is a decentralized training framework that aggregates many local models into a global model without sharing raw data, thereby complying with the requirements for data security \cite{10681457,10083185}. 

The standard approach for FL consists in training a local model at each client involved in the process, using the client's own data, and then broadcasting the parameters of local models to a global server. The server then combines these parameters to obtain a global model, the parameters of which are sent back to the clients. Repetition of this process for several rounds allows the global model to implicitly learn from the data of different clients, without having to directly share these data \cite{mcmahan2017communication}. Although FL models improve the privacy and security of DL systems trained from multiple sources, their performance and efficiency may be affected by two factors: \textbf{1)} \textcolor{black}{the communication and computation resources needed for global transmission and local training;} \textbf{2)} data distribution shifts among the clients. \textcolor{black}{For example, discrepancies in the distribution of data (i.e., data heterogeneity) can influence the performance of FL models, especially in medical imaging where the data acquisition process and patient demographics can vary largely from one site (e.g., hospital or clinic) to another \cite{10288131,chaddad2023enhancing}. This challenge can limit the generalizability of FL models where the test data are from different sources with large heterogeneity \cite{mora2024enhancing}.} Recent studies have shown that VLMs trained on natural images do not perform well on classification tasks involving medical images, highlighting the need for specialized adaptation mechanisms \cite{huix2024natural}. Moreover, the communication costs associated with parameter transmission can be prohibitive if the model is large, which poses a challenge for applications that require timely communication where high transmission latency is not allowed \cite{wu2024facmic}. This is especially problematic for VLMs such as \clip{} that have more than $10^8$ parameters (version based on ViT-B/32) \cite{lu2023fedclip}.

\textcolor{black}{This study investigates the potential of natural pretrained VLMs in the context of FL and propose a novel framework called Federated  Adversarial Adaptation of \clip{} (FAA-CLIP) addressing the challenges of data heterogeneity, communication, and computation costs.} FAA-CLIP handles the large size of CLIP with a light-weight feature adaptation module (FAM) that adapts this VLM to each client's data. Keeping the parameters of CLIP frozen, our method only updates and broadcasts the parameters of the clients' FAM, thereby reducing the overall computational and communication costs. Furthermore, compared to existing FL approaches, FAA-CLIP addresses the problem of data heterogenity in different clients directly, using a specialized domain adaptation (DA) module. The proposed module enables FAA-CLIP to learn domain-invariant representations with a domain classifier that predicts whether a given sample is from the local client or from a global reference data set. \textcolor{black}{The novelty of this study lies in our proposal of a training and communication efficient FL approach using a lightweight FAM and a domain adaptation module for adapting CLIP to both natural and medical datasets. Further calibration analysis conducted demonstrates the potential of FAA-CLIP for calibration tasks.}

The contributions of our work can be summarized as follows:
\begin{enumerate}
    % \item Although past studies have applied federated learning in medical imaging, to our knowledge, we are the first to explore FL on VLMs like CLIP in this context. 
    
    \item We propose a specialized feature adaptation module (FAM) for adapting CLIP to natural and medical imaging data from multiple sites, which significantly reduces the computational and communication costs of our solution. 
    
    \item Unlike existing apporoaches like FedCLIP \cite{lu2023fedclip}, our FAA-CLIP method addresses the problem of domain shifts directly using a domain adaptation (DA) module. As shown in our results, this DA module boosts performance in the various test scenarios.

    %\item We propose a novel feature attention module that entirely leverages the features extracted by the CLIP visual encoder and fine-tunes this module rather than the entire \clip{} to reduce resource costs.
    
    \item Through experiments on six publicly available classification benchmark datasets, including both natural and medical images, we demonstrate the higher classification accuracy and balanced accuracy of FAA-CLIP over state-of-the-art FL approaches.
\end{enumerate}

%The structure of this paper is summarized as follows. Section \ref{S:2} presents the related work on FL and FMs. In Section \ref{S:3}, we describe our model, while Section \ref{S:4} illustrates the experimental results. Finally, in Section \ref{S:5}, we provide our concluding remarks.

\section{Related Work}\label{S:2}

Before presenting our FAA-CLIP method, we give a brief overview of related work on federated learning and CLIP-based methods in the literature. 

\mypar{Federated learning (FL).} The increasing awareness of challenges related to data privacy and security has urged researchers to develop novel machine learning solutions that can learn from multiple sources of data without having to share this data \cite{voigt2017eu}. Among the many solutions proposed for this issue, FL has emerged as the most promising and popular approach \cite{lee2024federated}. Methods based on this approach can be roughly divided into three categories: horizontal federated learning (HFL), vertical federated learning (VFL), and federated transfer learning (FTL) \cite{10288131}. In HFL, clients share the same feature space while holding different samples. On the other hand, VFL in the FL setting where the datasets of individual clients contain information about the same samples but have different features. Last, in the FTL setting, client datasets differ both in terms of features and samples, with limited overlaps. 

One of the leading HFL methods, FedAVG \cite{mcmahan2017communication}, proposes a simple averaging mechanism to combine the parameters of locally-trained models into a single global one. Although it preserves data privacy by only sharing parameters, this method  may still suffer from the problem of data heterogeneity between clients \cite{sattler2019robust}. To solve this issue, FedProx added a proximal regularization term and allowed slight model gaps between clients and the server \cite{li2020federated}. In \cite{li2021fedbn}, authors argued that local distribution shifts in separate clients can be handled efficiently via batch normalization (BN) parameters. Based on this idea, they proposed a FedBN method that updates the BN parameters of each client locally without averaging them on the global server. On the other hand, MOON \cite{li2021model} used a contrastive loss between local clients and global servers to improve overall performance. 

Despite their improved performance, these methods mostly ignored the problem of generalization, which is also an important challenge in FL. To solve this problem, the work in \cite{yuan2021we} proposed a method to address separately the performance gaps arising from unseen client data (out-of-sample gap) and from unseen client distributions (participation gap). Likewise, the authors of \cite{qu2022generalized} introduced a FL framework based on the Sharpness Aware Minimization (SAM) local optimizer to improve the model's generalization ability. Moreover, the method presented in \cite{zhang2023federated} employed a new variance reduction regularizer to dynamically calibrating the aggregation weights, thereby improving generalization ability. \textcolor{black}{In \cite{qin2023fedapen}, they proposed a novel FL framework using mutual learning to provide a personalized local model with feasible performance. However, it introduces extra local training epochs.} Although these approaches boost the model robustness to distribution shifts, they were not intended for foundation models such as the one used in our work.

\mypar{Contrastive Language Image Pre-training (CLIP)} is an effective and scalable method learning representations jointly from image and text data \cite{radford2021learning}. The CLIP model, which was trained from scratch using over 400 million image-text data pairs collected from the Internet \cite{radford2021learning}, recently demonstrated remarkable performance in a wide range of image understanding tasks \cite{fan2024improving}. However, the integration and use of VLMs like CLIP in FL applications is stll in its infancy. In \cite{guo2023promptfl}, authors proposed replacing full VLM training with federated text prompt learning to reduce communication costs. Nevertheless, significant computational costs are still required to train the VLM. The work in \cite{singha2023ad} also highlighted the challenge of designing task-specific prompts for \clip{}. \textcolor{black}{Furthermore, in \cite{lu2023fedclip}, an adapter-based strategy was designed to leverage the power of pre-trained VLMs.} This strategy, which only tunes and aggregates the parameters of adapters, demonstrated its effectiveness on several natural datasets. Inspired by the superior performance of \clip{} for zero shot and few shot tasks, the authors of \cite{shi2024clip} used \clip{} to optimize FL between the server and client models,  alleviating the problem of data heterogeneity between clients. However, this approach incurred large communication costs as it requires sharing the entire model. \textcolor{black}{In \cite{zanella2024low}, they introduced low-rank adaptation (LoRA) for large language models to reduce computational costs, and this technique is widely used in the CLIP backbone in recent studies \cite{chen2024benchmarking}. In \cite{guo2023pfedprompt}, they proposed to use personalized prompts instead of global prompts for aggregation. The main idea is that they decoupled the parameters into base (global) and personalized (local) parameters.} \textcolor{black}{In \cite{chen2024feddat}, they proposed an adapter-based dual teacher technique using mutual knowledge distillation to adapt VLMs in FL tasks. However, it has not been validated in medical datasets.} Furthermore, in \cite{guo2023promptfl}, they proposed to aggregate the prompts instead of a shared global model to reduce communication costs (PromptFL). 

\textcolor{black}{Unlike previous studies, we explore the potential of natural pre-trained CLIP in the FL context for both natural and medical images with an acceptable resource cost. We also introduce calibration analysis \cite{tu2024closer} into CLIP-based FL approaches, an area that has been less explored in this context.}

\begin{figure*}
    \centering
    \includegraphics[width = 0.96 \textwidth]{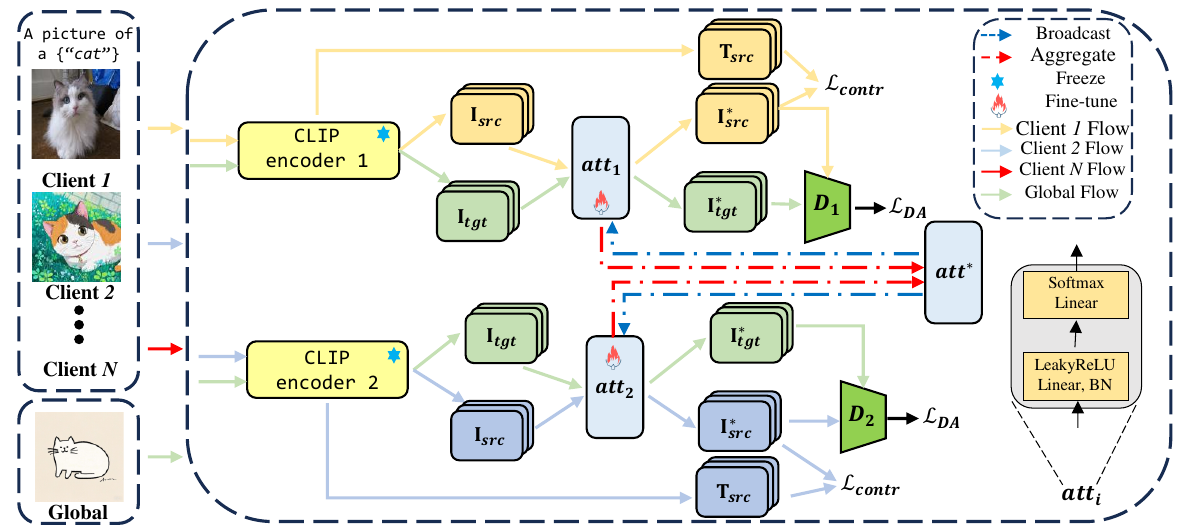}
    \caption{\textcolor{black}{FAA-CLIP pipeline. Each client trains its local model separately, optimizing only the parameters of its local feature adaptation module (FAM) ($att_i$) and domain classifier $D_i$ using contrastive and domain adaptation losses. After receiving the local client parameters, the server aggregates them into a global (FAM) ($att^{*}$) whose parameters are transmitted back to clients.}}
    \label{fig:framework}
\end{figure*}

%AC
\section{Methodology}\label{S:3} 

Before presenting our FAA-CLIP method for the federated adversarial adaptation of CLIP, we first introduce the FL setting considered in our study.

\subsection{Problem formulation}
In our FL setting, we work with a set of $N$ clients $\{C_1, C_2, \cdots, C_N\}$, each client $C_i$ having a private dataset $\data_i = \{(\xx_{i,j},\, y_{i,j})\}^{n_i}_{j=1}$. As in related studies \cite{lu2023fedclip}, we assume that the data of separate clients have the same input and output space, but follow different distributions, i.e., $P(\data_{i'}) \neq P(\data_{i})$, $\forall i' \neq i$. Each dataset $\data_i$ consists of three non-overlapping parts, namely a training $\data_i^{\mr{train}}\!=\!\{(\textbf{x}_{i,j}^{train},y_{i,j}^{train})\}_{j=1}^{n_i^{train}}$, a validation $\data_i^{\mr{val}}\!=\!\{(\textbf{x}_{i,j}^{val},y_{i,j}^{val})\}_{j=1}^{n_i^{val}}$ and a test set $\data_i^{\mr{test}}\!=\!\{(\textbf{x}_{i,j}^{test},y_{i,j}^{test})\}_{j=1}^{n_i^{test}}$. Our goal is to learn a global model $f_{\uptheta}(\cdot)$  while preserving data privacy and security. This model provides a good performance on the test data of every client, i.e.,
\begin{equation}
     \mathop {\min}\limits_f \ \frac{1}{N} \sum_{i=1}^N \frac{1}{n_i^{\mr{test}}} \sum_{j=1}^{n_i^{\mr{test}}} \ell(f_\uptheta(\xx^{\mr{test}}_{i,j}),\, y^{\mr{test}}_{i,j}),
\end{equation}
based on a given loss function $\ell$. For generalization, we assume that there exist $Q$ different clients $\{M_1,M_2,\cdots,M_Q\}$ with data $\data_i^M = \{(\xx_{i,j},\, y_{i,j})\}^{n_i}_{j=1}$. Our objective is for the model to perform on clients that \emph{were not included} in the local training phase, i.e.,
\begin{equation}
    \mathop {\min}\limits_f \ \frac{1}{M} \sum_{i=1}^M \frac{1}{m_i^{\mr{}}} \sum_{j=1}^{m_i^{\mr{}}} \ell(f_\uptheta(\xx^{\mr{}}_{i,j}),\, y^{\mr{}}_{i,j}).
\end{equation}

\subsection{Our FAA-\clip{} framework}

Our FL framework for \clip{}-based image classification, which is illustrated in Figure \ref{fig:framework}, comprises three key components: 1) a \emph{feature adaptation module} (FAM) for efficient updating and sharing of models, 2) a \emph{feature adaptation strategy} addressing the problem of data distribution shifts between clients, and 3) a \emph{global aggregation strategy} to combine the learned features from multiple clients. We present each of these components in the following.

\mypar{Feature Adaptation Module (FAM) training.} We use a pretrained \clip{} model, comprising an image encoder $e_I(\cdot)$ and a text encoder $e_T(\cdot)$, to extract features from the data for each client $C_i$. For a training example $\xx_j \in \data_{i}^{\mr{train}}$, we denote as $\fImg_j = e_I(\xx_j) \in \real^D$ the $D$-dimensional vector of image features. For text features, we use the standard prompt ``\texttt{a picture of a \{class\}}'' as input to the text encoder to obtain features $\fTxt_j = e_T(\xx_j) \in \real^D$. 

Pre-trained VLMs can extract a rich set of features, however, not all of those are suitable for learning a specific task. This is particularly true for detecting and classifying abnormal regions such as lesions in medical images, as these regions are absent in normal images and typically represent a small part of the image. \textcolor{black}{To identify the regions of focus for locally trained models, we introduce a local FAM, denoted as $att_i(\cdot)$. The main idea of FAM is to reduce the training costs while adapting the features extracted by CLIP to the new task through a series of nonlinear transformations. This FAM takes as input image features $\fImg$ and returns an attention mask $att_i(\fImg)\in [0,1]^D$. The attention mask is typically generated using the Softmax function. The mask is then used to generate the features of the masked images $\fImg^* = att_i(\fImg) \otimes \fImg$, where $\otimes$ is the Hadamard product (element-wise).}

We measure the probability that an example $\xx_j$ belongs to a class $c$ using the cosine similarity between the image features of $\xx_j$ and the text features $\fTxt_c$ corresponding to the prompt of $c$:
\begin{equation}\label{eq:CLIP-classif}
p(\mathrm{Y}\!=\!c \, | \, \xx_j) \, =\,  \frac{\exp(s_{j,c}/\tau)}{\sum_{c'=1}^K \exp(s_{j,c'}/\tau)}, \ \ \text{with }
s_{j,c} = \frac{\langle\fImg^{*}_j, \fTxt_c\rangle}{\|\fImg^{*}_j\| \!\cdot\!\|\fTxt_c\|}
\end{equation}
where $\tau$ is the Softmax temperature parameter. 

Keeping the image and text encoders frozen, we train the local FAMs by minimizing a contrastive loss $\loss_{\mr{contr}}$ that pushes together the image and text features from the same training example and pulls apart non-matching ones. Following \cite{radford2021learning}, we compute the contrastive loss over batches of size $B$. Let $\SSS$ be the $B\!\times\!B$ matrix where $s_{j,j'}$ is the cosine similarity between the image features $\fImg^{*}_j$ and $\fTxt_{j'}$ as measured in Eq (\ref{eq:CLIP-classif}). We compute an image probability matrix $\PP = \mr{softmax}(\SSS/\tau) \in [0,1]^{B \times B}$ and a text probability matrix $\QQ = \mr{softmax}(\tr{\SSS}/\tau) \in [0,1]^{B \times B}$. The contrastive loss is then formulated as follows:
\begin{equation}
\loss_{\mr{contr}} \, = \, -\frac{1}{B}\sum_{j=1}^B \frac{1}{2}\Big(\log p_{j,j} \, + \, \log q_{j,j}\Big). 
\end{equation}

\mypar{Feature adaptation.} We use a domain adversarial loss to align feature representations between clients \cite{ganin2016domain}. For each local client $i$, this loss trains a domain classifier (\emph{discriminator}) $D_i$ that predicts whether a given image representation $\fImg_j$ is from a \emph{source} ($D_i(\fImg_j) =1$) or a \emph{target} domain ($D_i(\fImg_j) =0$). For client $i$, the source domain contains the image features $\fImg_{j}$ of the client's training samples, i.e. $\xx_{j} \in \data_i$. Since private data cannot be shared in our FL scenario, we use a global set of unlabeled images $\data_{\mr{tgt}}$ to obtain the representations of the target domain. This requirement does not constitue a significant limitation as there are many publicly available datasets in medical imaging (e.g., UK Biobank \cite{sudlow2015uk} or Human Connectome Project -- HCP \cite{van2012human} for brain images). For each client $i$, batches of $2B$ samples are generated by randomly selecting the same number of source and target examples. Denoting as $z_j = \mathbbm{1}(\xx_j \in \data_i)$ the domain label of example $\xx_j$, the domain adaptation (DA) loss is then defined using cross-entropy, as follow:
\begin{equation}\label{eq:total_loss}
\begin{aligned}
   \mathcal{L}_{DA} \, = \, - \frac{1}{2B} \sum_{j=1}^{2B} z_j\log D_i(\fImg^*_j)
   \, + \, (1-z_j)\log \big(1-D_i(\fImg^*_j)\big)
\end{aligned}
\end{equation}
This loss is minimized for the domain classifier $D_i$, but maximized for the adapter $att_i$ so that the adapted image features of each client's data are indistinguishable from those on the shared reference data. Combined with the contrastive loss of Eq. (\ref{eq:CLIP-classif}), this gives rise to the following adversarial objective:
\begin{equation}\label{EQ:8}
\min_{att_i} \ \max_{D_i} \ \loss \, = \, \loss_{\mr{contr}}(att_i) \, - \, \lambda \loss_{\mr{DA}}(att_i, D_i),
\end{equation}
where $\lambda$ is a hyper-parameter controlling the trade-off between these two loss terms. In practice, this minimax optimization problem is converted to standard minimization using a gradient reversal layer \cite{ganin2016domain}.

\begin{algorithm}\small
\caption{Training procedure of FAA-CLIP.} 
\begin{algorithmic}[1]
        \Require $N$ clients' data $\{{\mathcal{D}_{i}}\}_{i=1}^{N}$, image encoder $e_{I}$, text encoder $e_{T}$, feature adaptation module $att^*$, domain classifier $D_i$
        \Ensure Feature adaptation module $att$
        \State For client $i$, compute the corresponding features obtain image and text features $I=e_{I}(image)$, $T=e_{T}(text)$
        \State For client $i$, train the local feature adaptation module, $att_i$ and domain classifier $D_i$ according to Eq. (\ref{eq:total_loss} );
        \State Send the current feature adaptation module $att_i$ to the server;
        \State Aggregate feature adaptation modules’ parameters via Eq. (\ref{EQ:9}) to obtain $att^*$;
        \State Transmit $att^*$ to each client $i$;
        \State \textbf{Repeat} steps 2 to 5 until convergence;
    \end{algorithmic}
    \label{Ag:2}
\end{algorithm}

\mypar{Global aggregation.} The last component of our FAA-CLIP framework is the aggregation strategy to combine the parameters of different clients into a single global model. This strategy works as follows. In every round, each client $C_i$ uploads its FAM parameters $\uptheta^{att}_i$ to the server. Thereafter, the server combines these parameters into a single vector $\uptheta^{att}_{\mr{global}}$ using a simple average
\begin{equation}\label{EQ:9}
    \uptheta^{att}_{\mr{global}} = \frac{1}{N}\sum_{i = 1}^N \uptheta^{att}_i.
\end{equation}
Subsequently, the server redistributes the global FAM parameters to every client. Compared to CLIP encoders, the FAM has less parameters. Hence our method has very low computational and communication costs. Algorithm \ref{Ag:2} summarizes the main steps of FAA-CLIP. 

% \textcolor{black}{Furthermore, as discussed in \cite{pan2009survey}, we evaluate how the method performs if the source domains contribute little. Thus, we reformulate the Eq. (\ref{EQ:9}) as follows.}
% \begin{equation}\label{EQ:10}
%     \uptheta^{\att}_{\mr{global}} = \sum\nolimits_{i = 1}^N \omega_i \cdot \uptheta^{att}_i, \ \
%     \omega_i = \frac{n_i^{\mr{train}}}{\sum\nolimits_{i' = 1}^N n_{i'}^{\mr{train}}}.
% \end{equation}

\section{Experiments}\label{S:4}

We start by presenting the data used for evaluating the proposed FAA-\clip{} model and then provide implementation details about our model's architecture, parameter optimization strategy and FL setup. Afterwards, we demonstrate our method's outstanding performance by comparing it to state-of-the-art FL approaches and baselines. Finally, we present several ablation studies to further investigate the impact of our method's different components and hyperparameters.

\subsection{Datasets}
Our method is evaluated on six datasets that contain both natural and medical images, which are described in the following.

\mypar{OfficeHome} is a large image classification benchmark, which contains 65 classes \cite{venkateswara2017deep}. This dataset has four sub-domains (Art, Clipart, Product, and Real World) with about 15,500 images. Following \cite{lu2023fedclip}, for each client, we divide the data into training (60\%), validation (20\%), and test (20\%) sets. In each experiment, three subdomains are used as clients (denoted as source), while one subdomain is considered as the global testing set (denoted as target).

\begin{figure}
    \centering
    \includegraphics[width=0.98\linewidth]{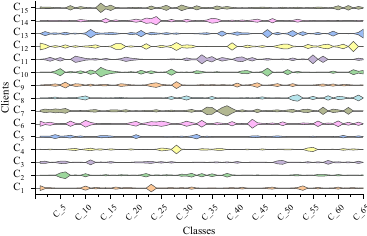}
    \caption{Example of data distribution in each client using kite graph in Multi-OF dataset. $C_1$ to $C_{15}$ indicate each client while $C\_1$ to $C\_65$ represent each class.}
    \label{fig:DataDistribution_OF}
\end{figure}

\mypar{Multi-OF} is a custom dataset derived from OfficeHome dataset. Following \cite{qin2023fedapen}, Dirichlet distribution is used to generate client data. Specifically, A, C and P domain are used for local training, while each domain is divided into five clients. For each domain data, we considered different $\alpha$ values used in the Dirichlet function to simulate heterogeneous data. Furthermore, domain R is used for the global evaluation following the study in \cite{lu2023fedclip}. Figure \ref{fig:DataDistribution_OF} shows the data distribution in each client using the kite graph (e.g., the size of polygon reflects the number of samples). As illustrated, there exist large heterogeneous data across all clients.

% \mypar{PACS} It is an object classification benchmark \cite{li2017deeper}. It has four sub-datasets, including photo, Art-painting, Cartoon, and Sketch. The data set has seven classes: dog, elephant, giraffe, guitar, horse, house, and person, and a total of 9991 images. Variations in image styles can be observed across these sub-datasets. Following OfficeHome, we view each sub-dataset as a client and choose three sub-datasets as sources while the rest served as the target client to evaluate. We evaluate our and other methods on PACS in a similar way to OfficeHome.

\mypar{ModernOffice31} is a refined version of the Office31 image classification benchmark \cite{Adaptiope} having four domains, Amazon (A), Webcam (W), DSLR (D) and Synthetic (S), with 31 classes in each domain. For every client, we divide the data into three parts: a training set (80\%), and a testing set (20\%). Following OfficeHome, we view each sub-dataset as a client and choose three sub-datasets as sources while the rest serves as the target client. We evaluate our and other methods on ModernOffice31 in a similar way to OfficeHome.

\mypar{Brain tumor (BT)} is a public Kaggle dataset, which has four different classes, namely glioma tumor, meningioma tumor, no tumor, and pituitary tumor\footnote{\url{https://www.kaggle.com/dsv/1183165}}. The training set is composed of 2,870 samples, while the test set has 394 samples. We randomly split the training set into three clients, and each client's data is then divided into two subsets: a training set (80\%), a validation set (10\%), and a testing set (10\%). Finally, we evaluated the global model using the original testing set.

\begin{table}\footnotesize
    \centering
    \renewcommand{\arraystretch}{1.2}
    \setlength{\tabcolsep}{1pt}
    \caption{Samples of each client.}
    \begin{tabular}{l|lcrcrcrcr}
    \toprule
     Dataset & \multicolumn{9}{c}{Clients}   \\
    \midrule  OfficeHome & & A & (2,427)\phantom{M} & C & (4,365)\phantom{M} & P & (4,439)\phantom{M} & R & (4,357)\phantom{M} \\
    ModernOffice31\phantom{M} & & A & (2,817)\phantom{M} & D & (498)\phantom{M} & S & (3,100)\phantom{M} & W & (795)\phantom{M}\\
    
    SC & & $C_1$ & (1,971)\phantom{M} & $C_2$ & (19,766)\phantom{M} & $C_3$ & (8,512)\phantom{M} & $C_{glo}$ & (6,233)\phantom{M} \\ 
    BT & & $C_1$ & (1,474)\phantom{M} & $C_2$ & (423)\phantom{M} & $C_3$ & (949)\phantom{M} & $C_{glo}$ & (394)\phantom{M}\\ 
HK & & $C_1$& (3,186)\phantom{M} & $C_2$ & (3,186)\phantom{M} & $C_3$ & (3,186)\phantom{M} & $C_{glo}$ & (1,104)\phantom{M}  \\
    \bottomrule
    \end{tabular}
    
    \label{tab:datasets}
\end{table}

\mypar{Multi-source skin cancer (SC).} We build this dataset from three sources: a kaggle public dataset, HAM10000, and ISIC2019 \cite{tschandl2018ham10000,codella2018skin,combalia2019bcn20000}. Since ISIC2019 lacks a test set, we divided it into two parts, one for training and the other for testing, with a ratio of 8:2. We selected the common classes for this dataset: actinic keratosis (AK), basal cell carcinoma (BCC), dermatofibroma (DF), melanoma (MEL), nevus (NV), pigmented benign keratosis (PBK), and vascular lesion (VL). For each client, we divide the data into two parts: a training set (80\%) and a testing set (20\%).

\mypar{HyperKvasir (HK).} To further evaluate our model, we use the largest gastrointestinal dataset \cite{Borgli2020}, comprising 10,662 labeled images with 23 classes. We randomly divide the dataset into four clients with the same number of samples, and we use three clients for training and employ the rest for testing. For each client in training, we split the data into two parts: a training set (70\%) and a testing set (30\%). 

A detailed description of the number of samples in each client, for all datasets, is provided in Table \ref{tab:datasets}.

%1474, 423, 949, 394
% \begin{figure*}[!ht]
%     % \centering
%     \begin{tabular}{cc}
%         \includegraphics[width = 0.47 \textwidth]{ecai-template/images/BT_test2.pdf} & \includegraphics[width = 0.47 \textwidth]{ecai-template/images/SC_test.pdf} \\
%     \end{tabular}
%     \vspace{-7pt}
%     \caption{Testing accuracy for each epoch using BT (\textbf{Left}) and SC (\textbf{Right}) datasets.}
%     \label{fig:BT_test}
% \end{figure*}

\begin{figure*}[!ht]
    \centering
    \begin{tabular}{c}

    \includegraphics[width=1\textwidth]{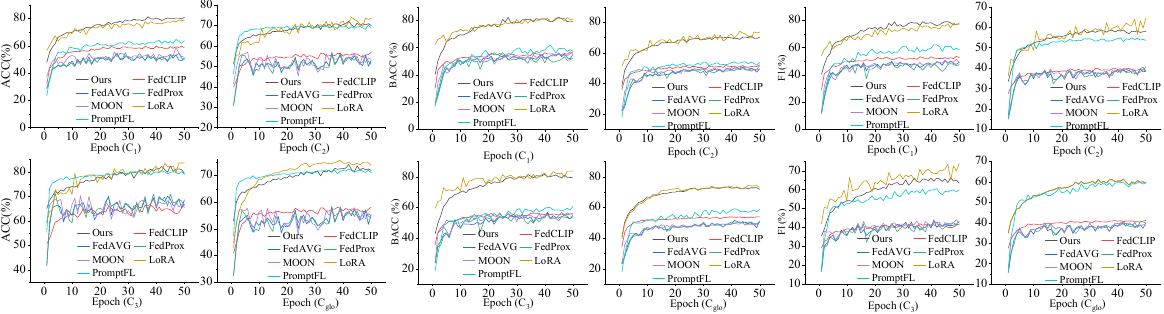}
    \end{tabular}
    \vspace{-3pt}
    \caption{\textcolor{black}{Testing accuracy (ACC), balanced accuracy (BACC) and macro-F1 for each communication epoch using skin cancer (SC) dataset for FAA-CLIP, FedProx, FedAVG, Fed\clip{}, LoRA$_{r=3}$, PromptFL and MOON.}}
    \label{fig:BT_test}
\end{figure*}

\begin{table}[!ht] \footnotesize
\caption{The computation overhead (all client's local training time per communication round) and the communication overhead (transmitted parameters per round). $\Sigma$, $\Sigma_{f}$, $\Sigma_{fc}$ and $\Sigma_{fam}$ are the parameter amount in the CLIP backbone, the CLIP without last layer, the fully-connection layer and the FAM, respectively. $\Sigma_{prompt}$ is the parameter amount in prompt learner.}
    \centering
    \renewcommand{\arraystretch}{1}
    \setlength{\tabcolsep}{3.5pt}
    
    \begin{tabular}{c|c|c|c|c|c}
    \toprule
    &\multicolumn{4}{c|}{Computation (mins)}&Communication  \\
    \midrule
    & OfficeHome & ModernOffice31&BT&SC&Param./round \\
    \midrule
    FedAVG&0.63 &1.01 &0.34 &2.06 &$2 \times \Sigma$ \\
    FedProx&0.75 &1.12 &0.39 &2.25 &$2 \times \Sigma$ \\
    MOON&0.74 &1.18 &0.41 & 2.21&$2 \times \Sigma$ \\
    LoRA$_{r=3}$&0.60 &0.98 &0.32 &2.03 &$2 \times 0.98\% \times \Sigma$  \\
    PromptFL&0.49 &0.87 &0.15 &2.05 & $2 \times \Sigma_{prompt}$  \\
    FedCLIP &0.44 &0.79 &0.16 &2.05 &$2 \times \Sigma_{fam}$ \\
\rowcolor{gray!15}    Ours&0.52 &0.93 &0.23 &2.06 &$2 \times \Sigma_{fam}$ \\

    \bottomrule
    \end{tabular}
    
    \label{tab:overhead}
\end{table}

\subsection{Implementation details} \label{SS:ID}
We adopted the pre-trained ViT-B/32 model as backbone for the image encoder in our \clip{}-based framework. Our light-weight FAM for adaptation is composed of five layers: a first linear layer, a batch normalization layer, a LeakyReLU layer, a second linear layer, and a softmax activation function. The domain classifier $D$ comprises eight layers: a first linear layer, a batch normalization layer, a ReLU layer, a second linear layer, a batch normalization layer, a ReLU layer, a third linear layer and a sigmoid activation function. During training, we keep the \clip{} encoders frozen and only optimize the parameters of the FAM and local domain classifier $D_i$ using the Adam optimizer with betas set to 0.9 and 0.98, a weight decay of 0.02, a fixed learning rate of 5$\times 10^{-5}$, an epsilon of 1$\times 10^{-6}$, and a batch size of 32 \cite{vaswani2017attention,kingma2014adam}. For all datasets, the hyperparameter $\lambda$ is set to 0.5 in Eq. (\ref{EQ:8}).

To have a comprehensive comparison, we include four recent FL approaches in our experiments: FedAVG \cite{mcmahan2017communication}, FedProx \cite{li2020federated}, MOON \cite{li2021model}, LoRA \cite{hu2021lora}, PromptFL \cite{guo2023promptfl} and Fed\clip{} \cite{lu2023fedclip}. The same experimental setting described above is used for all tested methods. In the case of FedProx, the parameter $\mu_{\mathrm{Prox}}$ is set to $1e^{-2} / 2$, while for MOON, we set $\mu_{\mathrm{\textsc{moon}}}$ as $1$. For LoRA, we set the low rank as 3 ($r=3$). Moreover, we fixed the random seed in PyTorch as $0$ to eliminate the variation caused by different seeds. For the OfficeHome dataset, the results of all baselines except MOON are based on \cite{lu2023fedclip}.

For FL, we set the number of global training rounds to 50 for all datasets. For each round, we perform a single epoch of local training and aggregate the parameters of all clients. For image pre-processing, we resized the images to $224 \times 224$, and normalized their intensity using z-score normalization for both the training and testing phases. \textcolor{black}{The testing environment used for the experiments is based on the Windows 11 operating system, and features an Intel 13900KF CPU with 128 GB of RAM and an RTX 4090 GPU. We use PyTorch 1.13.1 with Python 3.8. The scikit-learn library\footnote{\url{https://scikit-learn.org/stable/modules/model\_evaluation.html##classification-metrics}} was used to calcuate all metrics.} We evaluated performance using the top-1 classification accuracy (ACC) as the primary metric. Furthermore, for the SC, Dermnet and HK datasets, we also consider balanced accuracy (BACC) (average of recall obtained on each class) and macro F1 score (average of F1 obtained on each class) as metrics due to the class imbalance in these datasets. 

\begin{table}[!ht]
\centering\footnotesize
\renewcommand{\arraystretch}{1}
    \setlength{\tabcolsep}{8pt}
        \caption{Accuracy(\%) in the OfficeHome dataset. \textbf{Bold} means the best.}
        %\resizebox{\linewidth}{!}
        {  
    \begin{tabular}{l|ccc|c|c}
    \toprule
 & \multicolumn{3}{|c|}{Source} & Target & \multirow[b]{2}{*}{Avg}  \\
     \cmidrule(l{3pt}r{3pt}){2-4} \cmidrule(l{3pt}r{3pt}){5-5}
   & A & C & P & R & \\
    \midrule 
    Fed\clip{} & 78.35 & 68.38 & 87.94 & 87.79 & 80.61 \\
    FedProx & 70.93 & 68.73 & 77.73 & 75.42 & 73.20 \\
    FedAVG & 70.93 & 68.73 & 77.73 & 75.42 & 73.20 \\
    MOON & 67.42 & 69.30 & 76.32 & 75.81 & 72.21  \\
    LoRA$_{r=3}$&73.75&\textbf{83.80}&90.62&83.80&82.99 \\
    PromptFL&77.50&83.22&\textbf{90.97}&85.48&\textbf{84.29} \\
    % Ours4&78.12&82.87&91.44&86.14&84.64 \\
    % Ours3&78.96&82.06&91.55&86.53&84.78\\
    % Ours2&78.33&82.87&92.13&84.81&84.53 \\
    \ccol Ours & \ccol \textbf{78.76} & \ccol {76.98} & \ccol {90.64} & \ccol \textbf{88.11} & \ccol {83.62} \\
   % Ours$^\Delta$ & 80.41 & 74.45 & 90.30 & 88.13 & 83.32\\
    \midrule
 & C & P & R & A & \\
    \midrule
   Fed\clip{} & 68.61 & 87.37 & 88.06 & 78.00 & 80.51 \\
    FedProx & 64.38 & 79.14 & 78.76 & 65.60 & 71.97 \\
    FedAVG & 64.38 & 79.14 & 78.76 & 65.60 & 71.97 \\
    MOON & 69.87 & 78.02 & 79.56 & 65.60 & 73.26 \\
    LoRA$_{r=3}$&\textbf{84.26}&90.74&86.34&69.96&82.83 \\
    PromptFL&{82.75}&\textbf{92.13}&87.73&71.88&83.62 \\
  \ccol Ours & \ccol {76.98} & \ccol {90.87} & \ccol \textbf{88.86} & \ccol \textbf{78.20} & \ccol \textbf{84.49} \\
  % Ours$^\Delta$ & 72.71 & 90.52 & 89.89 & 78.20 & 82.83 \\
    \midrule
 & A & P & R & C & \\
    \midrule
    Fed\clip{} & 78.97 & 87.60 & 87.60 & 63.69 & 79.46 \\
    FedProx & 73.81 & 80.38 & 80.48 & 57.64 & 73.08 \\
    FedAVG & 73.81 & 80.38 & 80.48 & 57.64 & 73.08 \\
    MOON & 70.93 & 80.38 & 80.83 & 52.94 & 71.27 \\
    LoRA$_{r=3}$&76.04&90.51&86.57&62.71&78.96 \\
    PromptFL&78.75&\textbf{92.71}&87.27&\textbf{66.66}&81.35 \\
 \ccol Ours & \ccol \textbf{81.44} & \ccol {91.65} & \ccol \textbf{90.58} & \ccol {66.39} & \ccol \textbf{82.51} \\
  % Ours$^\Delta$ & 78.92 & 90.29 & 89.19 & \textbf{66.64} & 81.26 \\
   \midrule
 & A & C & R & P & \\
    \midrule
    Fed\clip{} & 78.56 & 68.50 & 87.37 & 87.52 & 80.49 \\
    FedProx & 69.07 & 66.21 & 77.79 & 71.64 & 71.18\\
    FedAVG & 69.07 & 66.21 & 77.79 & 71.64 & 71.18 \\
    MOON & 71.55 & 67.70 & 79.33 & 71.21 & 72.45 \\
    LoRA$_{r=3}$&75.62&\textbf{84.49}&86.00&85.42&82.88 \\
    PromptFL&77.71&80.79&88.31&85.19&83.0  \\
   \ccol Ours & \ccol \textbf{81.03} & \ccol {75.49} & \ccol \textbf{90.81} & \ccol \textbf{89.34} & \ccol \textbf{84.29} \\
% Ours$^\Delta$ & 79.34 & 74.08 & 88.96 & 89.21 & 82.90 \\
    % ModernOffice31
    \bottomrule
    \end{tabular}
    }

    \label{tab:OfficeHome}
% {Ours$^\Delta$ indicates the results using data augmentation.}
\end{table}

\begin{table}[!ht]
\centering\footnotesize
    \setlength{\tabcolsep}{8pt}
        \caption{Accuracy(\%) in the ModernOffice31 dataset. \textbf{Bold} means the best.}
        %\resizebox{\linewidth}{!}
        {  
    \begin{tabular}{l|ccc|c|c}
    \toprule
 & \multicolumn{3}{|c|}{Source} & Target & \multirow[b]{2}{*}{Avg}  \\
     \cmidrule(l{3pt}r{3pt}){2-4} \cmidrule(l{3pt}r{3pt}){5-5}
    & A & D & S & W & \\
    \midrule 
   Fed\clip{} & 91.82 & 94.94 & 65.00 & 88.42 & 85.04 \\
    FedProx & 82.41 & 81.82 & 74.03 & 79.49 & 79.43 \\
    FedAVG & 59.07 & 69.38 & 59.67 & 64.53 & 63.16 \\
    MOON & 84.01 & 81.82 & 74.68 & 83.14 & 80.91 \\
    LoRA$_{r=3}$&94.32&94.95&\textbf{93.23}&91.54&\textbf{93.51} \\
    PromptFL&90.05&78.79&79.52&68.23&79.15\\
  \ccol Ours & \ccol \textbf{95.38} & \ccol \textbf{98.99} & \ccol {82.74} & \ccol \textbf{91.95} & \ccol {92.26} \\
    \midrule
 & D & S & W & A & \\
    \midrule
   Fed\clip{} & 95.95 & 65.32 & 88.05 & 90.06 & 84.84 \\
    FedProx & 89.90 & 59.52 & 90.57 & 66.88 & 76.71 \\
    FedAVG & 88.89 & 49.19 & 91.19 & 42.81 & 68.02 \\
    MOON & 95.96 & 65.81 & 91.19 & 66.92 & 79.97 \\
    LoRA$_{r=3}$&97.98&\textbf{93.87}&\textbf{98.74}&76.28&\textbf{91.72} \\
    PromptFL&81.82&76.13&83.65&68.93&77.63\\
    \ccol Ours & \ccol \textbf{100} & \ccol {80.32} & \ccol {94.34} & \ccol \textbf{92.19} & \ccol { 91.71}\\
    
    \midrule
 & S & W & A & D & \\
    \midrule
    Fed\clip{} & 64.67 & 86.16 & 91.82 & 89.15 & 82.95 \\
    FedProx & 73.55 & 90.57 & 81.53 & 89.36 & 83.75 \\
    FedAVG & 76.94 & 93.08 & 82.59 & 92.17 & 86.19 \\
    MOON & 75.00 & 92.45 & 87.39 & 85.74 & 85.14 \\
    LoRA$_{r=3}$&\textbf{92.58}&\textbf{96.23}&93.43&93.54&\textbf{93.95} \\
    PromptFL&79.68&77.99&90.41&67.71&78.95\\
    \ccol Ours & \ccol {81.93} & \ccol {93.71} & \ccol \textbf{95.38} & \ccol \textbf{95.98} & \ccol {91.75} \\
    
   \midrule
 & W & A & D & S & \\
    \midrule
    Fed\clip{} & 87.42 & 91.47 & 95.95 & 54.77 & 82.40 \\
    FedProx & 91.82 & 81.17 & 89.90 & 46.45 & 77.33 \\
    FedAVG & 95.60 & 80.46 & 97.98 & 45.87 & 79.97 \\
    MOON & 94.34 & 82.06 & 89.90 & 54.77 & 80.27 \\ 
    LoRA$_{r=3}$&\textbf{97.48}&\textbf{95.56}&\textbf{98.99}&48.93&85.24 \\
    PromptFL&81.76&88.45&85.86&32.49&72.14\\
   \ccol Ours & \ccol {96.23} & \ccol {94.85} & \ccol \textbf{98.99} & \ccol \textbf{57.10} & \ccol \textbf{86.79 }\\
    \bottomrule
    \end{tabular}
    }
\label{tab:ModernOffice31}
\end{table}

\begin{figure}[]
    % \centering
    \includegraphics[width = 0.98\linewidth]{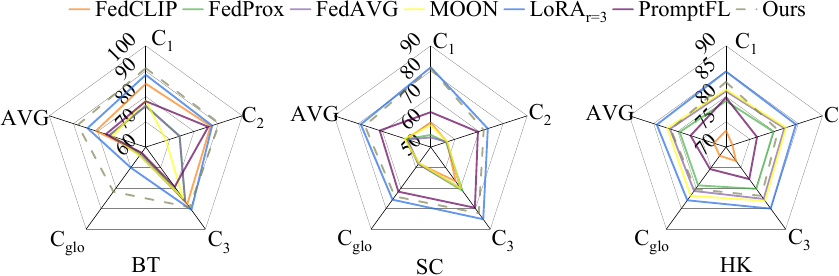}
    % \vspace{-3pt}
    \caption{\textcolor{black}{Testing accuracy of baselines and FAA-CLIP for BT, SC and HK datasets.}}
    \label{fig:acc_medical_all}
\end{figure}

\begin{table}[!ht]
    \setlength{\tabcolsep}{4pt}
        \caption{BACC (\%) and F1 score (\%) in SC and HK datasets. \textbf{Bold} means the best.}
        \resizebox{\linewidth}{!}{  
    \begin{tabular}{ll|cc|cc|cc|cc|cc}
    \toprule
 & & \multicolumn{6}{|c|}{Source} & \multicolumn{2}{c}{Target} & \multicolumn{2}{|c}{\multirow[b]{2}{*}{Avg}}  \\
     \cmidrule(l{3pt}r{3pt}){3-8} \cmidrule(l{3pt}r{3pt}){9-10}
 & & \multicolumn{2}{c|}{$C_1$} & \multicolumn{2}{c|}{$C_2$ } & \multicolumn{2}{c|}{$C_3$ } & \multicolumn{2}{c|}{$C_{glo}$} & \\
  \midrule
       %\multicolumn{11}{c}{\textsc{SC}} \\
     % \midrule
    %\midrule 
    \parbox[t]{2mm}{\multirow{7}{*}{\rotatebox[origin=c]{90}{SC}}}
 & Fed\clip{} & 56.79 & 53.52 & 51.98 & 40.21 & 56.33 & 41.91 & 53.74 & 41.09 & 54.71 & 44.18 \\
 & FedProx & 54.46 & 50.09 & 49.51 & 40.29 & 55.75 & 42.41 & 50.63 & 40.50 & 52.59 & 43.32 \\
 & FedAVG & 55.93 & 50.38 & 50.61 & 39.33 & 56.82 & 43.95 & 51.38 & 39.97 & 53.68 & 43.41 \\
 & MOON & 56.22 & 51.38 & 51.79 & 39.84 & 56.73 & 43.97 & 50.78 & 39.75 & 53.88 & 43.73   \\
 
 &LoRA$_{r=3}$&\textbf{83.73}&\textbf{81.34}&\textbf{72.05}&\textbf{63.98}&\textbf{81.71}&\textbf{71.96}&\textbf{74.23}&\textbf{63.54}&\textbf{77.93}&\textbf{70.2}  \\
 
 &PromptFL&60.43&61.98&53.93&54.72&59.94&59.91&59.88&59.50&58.55&59.03   \\
 & \ccol Ours & \ccol {82.14} & \ccol {79.01} & \ccol {70.95} & \ccol {58.60} & \ccol {81.63} & \ccol {66.35} & \ccol {73.15} & \ccol {61.22} & \ccol {76.97} & \ccol {66.29} \\
\midrule
    \parbox[t]{2mm}{\multirow{7}{*}{\rotatebox[origin=c]{90}{HK}}}
 & Fed\clip{} & 50.86 & 45.46 & 53.03 & 45.26 & 50.63 & 45.04 & 49.62 & 46.73 & 51.03 & 45.62 \\
 & FedProx & 57.13 & 51.06 & 56.24 & 52.22 & 57.35 & {53.09} & 52.66 & 48.33 & 55.84 & 51.17  \\
 & FedAVG & 56.05 & 50.63 & 57.33 & 51.71 & 56.51 & 52.83 & 53.01 & 50.11 & 55.72 & 51.32  \\
 & MOON & 55.94 & 51.82 & 56.53 & {52.68} & {57.70} & 52.55 & {54.65} & {50.73} & 56.20 & {51.94}  \\
 
 &LoRA$_{r=3}$&60.08&\textbf{57.29}&\textbf{60.45}&\textbf{56.20}&\textbf{59.14}&\textbf{56.90}&\textbf{56.23}&\textbf{55.17}&\textbf{58.98}&\textbf{56.39}  \\

 &PromptFL&48.35&47.78&45.70&44.39&47.25&45.72&44.74&42.77&46.51&45.16   \\
 & \ccol Ours & \ccol \textbf{62.40} & \ccol {53.59} & \ccol {58.02} & \ccol 50.49 & \ccol 57.38 & \ccol 50.51 & \ccol 53.48 & \ccol 49.51 & \ccol {57.82} & \ccol 51.02  \\
 
    \bottomrule
    \end{tabular}
    }
    \label{tab:SCBACC}
\end{table}

\begin{table*}[]\footnotesize
    \centering
    \renewcommand{\arraystretch}{1}
    \setlength{\tabcolsep}{4.7pt}
    \caption{Accuracy (\%) of test samples of OfficeHome based on the Dirichlet distribution (non-IID). The client domains are A ($C_1$ to $C_5$), C ($C_6$ to $C_{10}$) and P ($C_{11}$ to $C_{15}$), while the global domain is R. \textbf{Bold} means the best.}
    \begin{tabular}
    {c|ccccc|ccccc|ccccc|c|c}
        \toprule
         &\multicolumn{15}{c|}{Source}&Target&\multirow{2}{*}{AVG} \\
         \cmidrule(l{3pt}r{3pt}){2-16} \cmidrule(l{3pt}r{3pt}){17-17}
        &$C_1$ &$C_2$ &$C_3$ &$C_4$ &$C_5$ &$C_6$ &$C_7$ &$C_8$ &$C_9$ &$C_{10}$ &$C_{11}$ &$C_{12}$ &$C_{13}$ &$C_{14}$&$C_{15}$&R & \\
        \midrule
        FedAVG&72.92&83.33&80.21&81.25&75.00&82.81&84.38&80.21&\textbf{88.28}&82.50&88.75&89.06&\textbf{92.50}&83.59&92.97&83.48&83.83 \\
        FedProx&73.96&\textbf{85.42}&\textbf{81.25}&\textbf{83.33}&\textbf{78.12}&\textbf{83.85}&\textbf{85.42}&\textbf{85.42}&\textbf{88.28}&\textbf{85.62}&88.75&90.10&91.88&85.16&92.19&83.36&\textbf{85.13} \\
        MOON&\textbf{76.04}&83.33&79.17&\textbf{83.33}&75.00&81.77&\textbf{85.42}&79.17&86.72&83.12&88.12&86.98&91.25&85.94&95.31&83.07&83.98 \\
        LoRA$_{r=3}$&65.62&\textbf{85.42}&72.92&77.08&67.19&80.73&83.85&78.12&\textbf{88.28}&79.38&88.75&94.79&91.88&84.38&\textbf{96.09}&81.23&82.23 \\
        PromptFL&53.12&82.29&71.88&69.79&60.94&69.27&67.19&61.46&81.25&74.38&86.88&84.38&88.12&78.91&87.50&44.55&72.62\\
        Fed\clip{}&71.88&83.33&77.08&80.21&75.00&69.79&71.35&67.71&78.12&73.75&86.88&86.98&\textbf{92.50}&78.91&93.75&87.04&79.64 \\
        \rowcolor{gray!15}Ours&71.88&\textbf{85.42}&78.12&\textbf{83.33}&76.56&77.60&74.48&70.83&80.47&75.62&\textbf{90.62}&\textbf{91.15}&\textbf{92.50}&\textbf{86.72}&\textbf{96.09}&\textbf{87.52}&82.43\\
        
        \bottomrule
    \end{tabular}
    
    \label{tab:MultiOF}
\end{table*}

\subsection{Results}\label{SS:Results}

\mypar{Communication and computational costs.} We record the time cost of each round for each method and the communication overhead in each round on OfficeHome (\{ACPR\}), ModernOffice31 (\{ADSW\}), BT and SC, as reported in Table \ref{tab:overhead}. As illustrated, FAA-\clip{} is training efficient compared to traditional FL techniques such as FedAVG, FedProx, and MOON, while it indicates lower computational costs compared to LoRA. Furthermore, the communication costs for our approach are comparable with Fed\clip{} ($\sim 5 \times 10^{5}$), while is $3 \times$ lower than LoRA ($\sim 1.5 \times 10^{6}$).

\mypar{Results for OfficeHome.} Table \ref{tab:OfficeHome} reports the best testing accuracy, including baselines and our method for the OfficeHome dataset. As can be seen, our method achieved the best average performance with an improvement of 1.29\% (target is R) compared with the second-best method (PromptFL). Improvements are particularly notable for domain R. 

\mypar{Results for Multi-OF.} Table \ref{tab:MultiOF} reports the accuracy of the test sets for the baselines and our method. As illustrated, fine-tuning the entire CLIP model can achieve feasible test ACC in local clients (e.g., 83.85\% ACC using FedProx in $C_6$), while indicates lower ACC in unseen domain (e.g., 83.48\% accuracy in domain R using FedAVG). Concerning training efficient methods such as LoRA, PromptFL, FedCLIP and ours, the use of deeper FAM and DA in FAA-\clip{} can lead to a higher ACC test in domain R (87.52\%) compared to LoRA (81.23\%) and PromptFL (44.55\%). Surprisingly, FAA-\clip{} provides the highest test ACC in $C_{11}$ to $C_{15}$, demonstrating the potential of FAA-\clip{} in FL.

\mypar{Results for ModernOffice31.} As reported in Table \ref{tab:ModernOffice31}, FAA-\clip{} also achieves comparable performance compared to LoRA in \{ADSW\} (92.26\% Avg vs 93.51\% Avg) and \{DSWA\} (91.71\% Avg vs 91.72\% Avg), while indicates the highest average test ACC (86.79\%) in \{WADS\} for the ModernOffice31 dataset. Notably, FAA-\clip{} gives remarkable performance in domain S, showing its potential for synthetic data.

\mypar{Results for BT.} Once more, our method obtains the best average performance (88.18\%) for the BT dataset (Figure \ref{fig:acc_medical_all}). In particular, our method reaches a test ACC above 80\% in $C_{glo}$, outperforming the second-best FL approach by over $\sim 10\%$. 

% The test accuracy for each epoch, shown in Figure \ref{fig:BTACC_Base}, also demonstrates the faster convergence of our method compared to other baselines.

% \begin{figure}[!ht]
%     \centering
%     \includegraphics[width = 0.48\textwidth]{FedCLIP+DA/TestImage/TMI_BT_ACC.pdf}
%     \caption{Testing accuracy (ACC) of each communication round for Ours, Fed\clip{}, FedAVG, FedProx and MOON in BT dataset.}
%     \label{fig:BTACC_Base}
% \end{figure}

\mypar{Results for SC.} The performance of the compared approaches for the SC dataset is summarized in Figure \ref{fig:acc_medical_all} and Table \ref{tab:SCBACC}. For ACC and BACC, FAA-\clip{} outperforms Fed\clip{} by over $10\%$. Furthermore, our method achieves an F1 score near 61\%, largely outperforming approaches like Fed\clip{} which only reach about 41\%. Moreover, a prompt-tuning technique such as PromptFL indicates less effective compared to FAA-\clip{} in skin cancer classification (e.g., 71.18\% AVG vs 77.37\% AVG for ACC). In addition, FAA-\clip{} can provide similar performance compared to LoRA with less communication costs as reported in Table \ref{tab:overhead}. Figure \ref{fig:BT_test} shows the test ACC, BACC and macro F1 for different communication epochs in each client. As illustrated, FAA-\clip{} achieves a fast convergence, reaching a high classification ACC in a few epochs.

\mypar{Results for HK.} As reported in Figure \ref{fig:acc_medical_all} and Table \ref{tab:SCBACC}, our method achieves an ACC, BACC and macro F1 score comparable to MOON and FedProx for this dataset, while having lower resource costs than these approaches. Our approach outperforms Fed\clip{} by a large margin ($>$8\%) in ACC. However, due to the class imbalance in the HK data set, FAA-\clip{} indicates a lower performance (e.g., BACC) compared to LoRA (57.82\% Avg vs 58.98\% Avg).

% Figure \ref{fig:BT_test} illustrates the test metrics for each epoch. As illustrated, our method is more robust compared to other approaches.

\begin{figure*}[ht]
    % \centering
    \setlength{\tabcolsep}{8pt}
    \begin{tabular}{cc}
\includegraphics[height = 0.38 \textwidth]{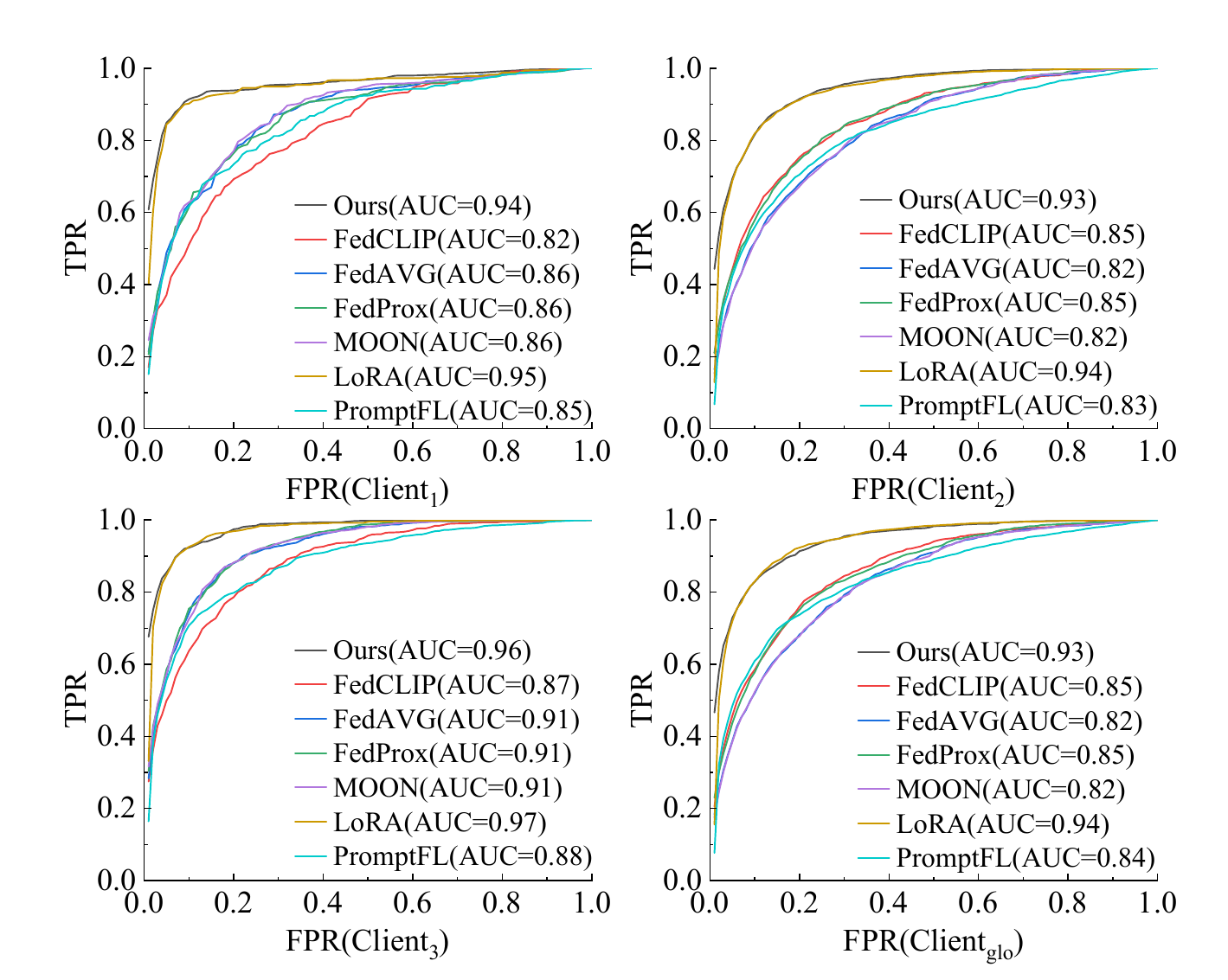} & 
\includegraphics[height = 0.38\textwidth]{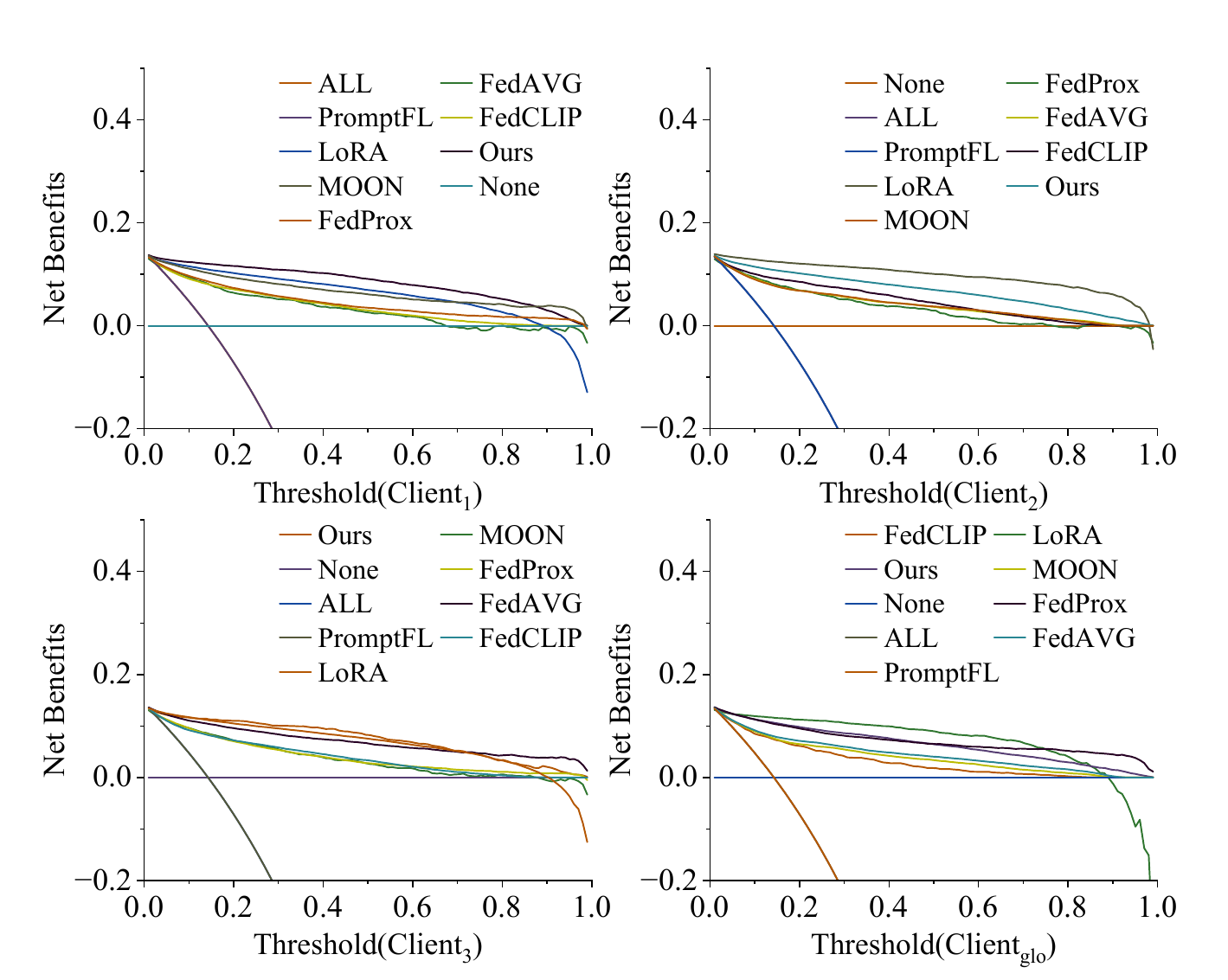} \\
    \end{tabular}
    \vspace{-3pt}
    \caption{\textcolor{black}{The ROC curves and DCA of FAA-CLIP, Fed\clip{}, FedAVG, FedProx and MOON in skin cancer (SC) dataset. The ROC can measure the performance of classifiers, while DCA can assess the net benefit for clinical practices. The \textbf{left} part is the ROC curves, while the \textbf{right} part is the DCA.}} 
    \label{fig:DCA_ALL}
\end{figure*}

\begin{table}
\centering\footnotesize
    \setlength{\tabcolsep}{7.7pt}        \caption{Accuracy (\%) in OfficeHome, BT and SC. \textbf{Bold} means the best.}
    \begin{tabular}{l|ccc|c|c}
    \toprule
      & \multicolumn{3}{|c|}{Source} & Target & \multirow[b]{2}{*}{Avg}  \\
     \cmidrule(l{3pt}r{3pt}){1-1}\cmidrule(l{3pt}r{3pt}){2-4} \cmidrule(l{3pt}r{3pt}){5-5}
   OfficeHome & A & C & P & R & \\
    \midrule 
       Ours & {78.76} & {76.98} & {90.64} & {88.11} & {83.62} \\
    ~~ w/ deeper FAM & \textbf{80.00} & \textbf{79.04} & {90.98} & 88.09 & \textbf{84.53} \\
    ~~ w/o shared BN & 78.76 & 76.97 & 90.64 & 88.11 & 83.62 \\
    % Ours$^\lambda$ & 78.14 & 73.65 & 89.96 & 87.88 & 82.41 \\
    % Ours$^\Lambda$ & 78.14 & 72.74 & \textbf{91.32} & 88.02 & 82.55 \\
    ~~ w/ shared $D$ & 78.97 & 76.74 & 90.64 & \textbf{88.16} & 83.63 \\
    \midrule
  BT & $C_1$ & $C_2$ & $C_3$ & $C_{glo}$ & \\
    \midrule
     Ours & \textbf{91.16} & \textbf{90.47} & \textbf{89.36} & \textbf{81.73} & \textbf{88.18} \\
    ~~ w/ deeper FAM & 87.07 & 89.28 & 89.94 & 79.44 & 86.43 \\
    ~~ w/o shared BN & 85.71 & 89.28 & 85.18 & 78.42 & 84.65 \\
    ~~ w/ shared $D$ & {90.47} & \textbf{90.47} & \textbf{89.36} & 80.20 & {87.62} \\
    \midrule
SC & $C_1$ & $C_2$ & $C_3$ & $C_{glo}$ & \\
\midrule
% Ours$^@$ & 84.52 & 77.69 & 89.24 & 81.64 & 83.27 \\
%     Ours$^{@\dag}$ & 85.53 & 78.39 & 88.78 & 81.20 & 83.47 \\
    ~~ w/o DA & 79.94 & 71.49 & 81.02 & 71.50 & 75.99  \\
    \bottomrule
    \end{tabular}\\
%{~w new FAM indicates the results using the new FAM, Ours$^*$ represents the results without sharing batch normalization layers parameters, and Ours$^\nabla$ are the results with aggregation for domain classifier $D$, Ours$^{@*}$ indicates the result without DA.}
    \label{tab:MoreAdap}
\end{table}

\begin{table*}[!ht]\footnotesize
    \setlength{\tabcolsep}{5.5pt}
        \caption{Accuracy (\%) in OfficeHome and BT dataset. \textbf{Bold} means the best.}
        \resizebox{\linewidth}{!}{  
    \begin{tabular}{c|ccccc|cccc|cccc|cccc}
    \toprule
 & \multicolumn{5}{c}{Attention module layer} & \multicolumn{4}{|c}{OfficeHome} & \multicolumn{4}{|c}{BT} & \multicolumn{4}{|c}{SC} \\
     \cmidrule(l{3pt}r{3pt}){2-6} \cmidrule(l{3pt}r{3pt}){7-10} \cmidrule(l{3pt}r{3pt}){11-14}\cmidrule(l{3pt}r{3pt}){15-18}
 & L1 & L2 & L3 & L4 & L5 & A & C & P & R & $C_1$ & $C_2$ & $C_3$ & $C_{glo}$ & $C_1$ & $C_2$ & $C_3$ & $C_{glo}$ \\
    \midrule 
   fea$_1$ & \checkmark & \ding{55} & \ding{55} & \ding{55} & \checkmark & 79.38 & 73.38 & 89.51 & 87.35 & 77.51 & 90.47 & 81.91 & 64.21 & 62.43 & 57.35 & 66.62 & 59.21 \\
   fea$_2$ & \checkmark & \checkmark & \ding{55} & \ding{55} & \checkmark & 78.96 & 78.69 & 92.78 & \textbf{88.50} & 87.75 & 88.09 & 87.23 & 80.71 & 80.96 & 70.60 & 81.19 & 73.01 \\
   fea$_3$ & \checkmark & \checkmark & \checkmark & \ding{55} & \checkmark & 79.38 & \textbf{81.09} & \textbf{91.77} & 87.33 & 86.39 & 85.71 & 85.11 & 80.96 & 79.94 & 71.64 & 81.96 & \textbf{74.15}  \\
   fea$_4$ & \checkmark & \ding{55} & \checkmark & \checkmark & \checkmark & \textbf{80.20} & 73.08 & 89.74 & 88.38 & 83.67 & \textbf{92.85} & 82.97 & 63.45 & 59.39 & 56.76 & 66.45 & 58.24  \\   fea$_5$ & \checkmark & \checkmark & \checkmark & \checkmark & \checkmark & 78.76 & 76.98 & 90.64 & 88.11 & \textbf{91.16} & 90.47 & \textbf{89.36} & \textbf{81.73} & \textbf{81.22} & \textbf{71.95} & \textbf{82.26} & {74.07 } \\
    \bottomrule
    \end{tabular}
    }
{
\\[-1pt]
\footnotesize
L1: Linear layer; L2: Batch normalization layer; L3: LeakyReLU layer; L4: Linear layer; L5: Softmax layer.}
    \label{tab:attn}
\end{table*}

\mypar{AUC-ROC analysis.} Figure \ref{fig:DCA_ALL} shows the receiver operator characteristics (ROC) curve with corresponding area under the curve (AUC) values for SC dataset. For simplicity, we used the one-vs-all method to measure the ROC and AUC and report the averaging curves/values for all classes. These curves show the more robust performance of our method for the medical datasets used in experiments.

\mypar{Decision curve analysis.} We further validate our method for clinical practice using a decision curve analysis (DCA) \cite{vickers2006decision} on SC dataset. Typically, the best prediction to maximize net benefit is the one with the highest curve at any given threshold probability. Our experimental datasets are multi-class, thus we verify our method against existing FL methods using averaged net benefits (the average value of each class). As can be seen in Figure \ref{fig:DCA_ALL}, our method provides a greater  net benefit than other models, especially in the Client$_1$. This finding is consistent with the higher AUC-ROC values obtained by our approach.

\begin{figure}
    \centering
    \includegraphics[width=0.49\linewidth]{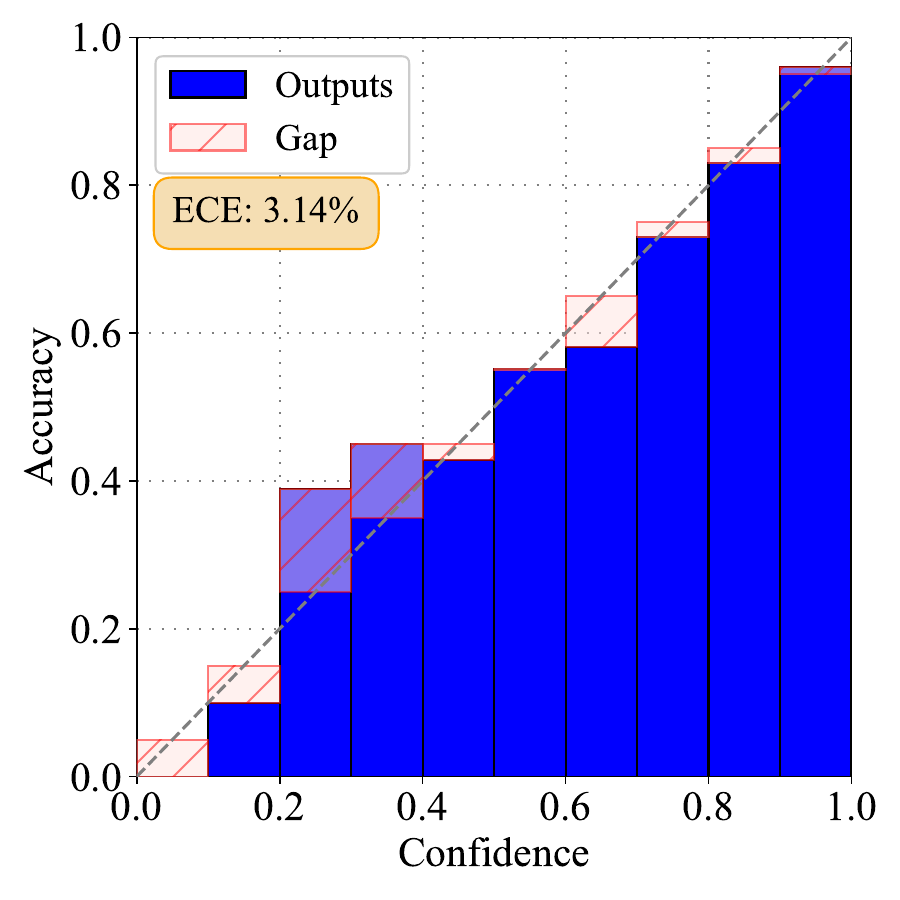} \includegraphics[width=0.49\linewidth]{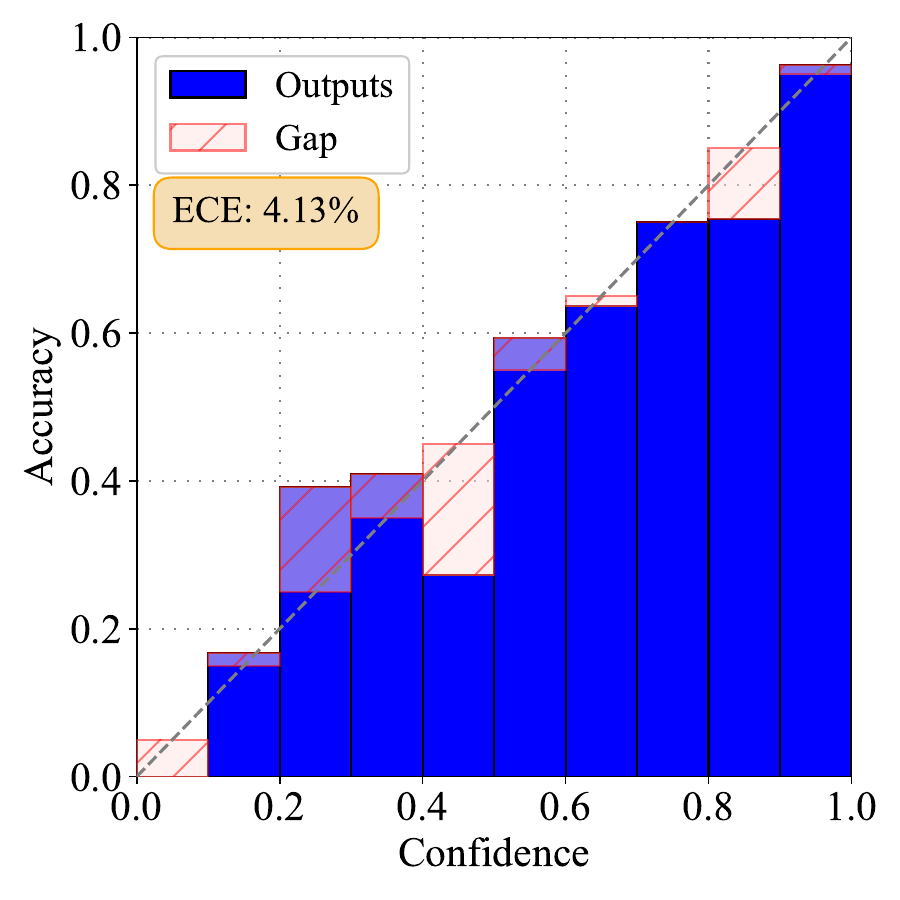}\\
    \includegraphics[width=0.49\linewidth]{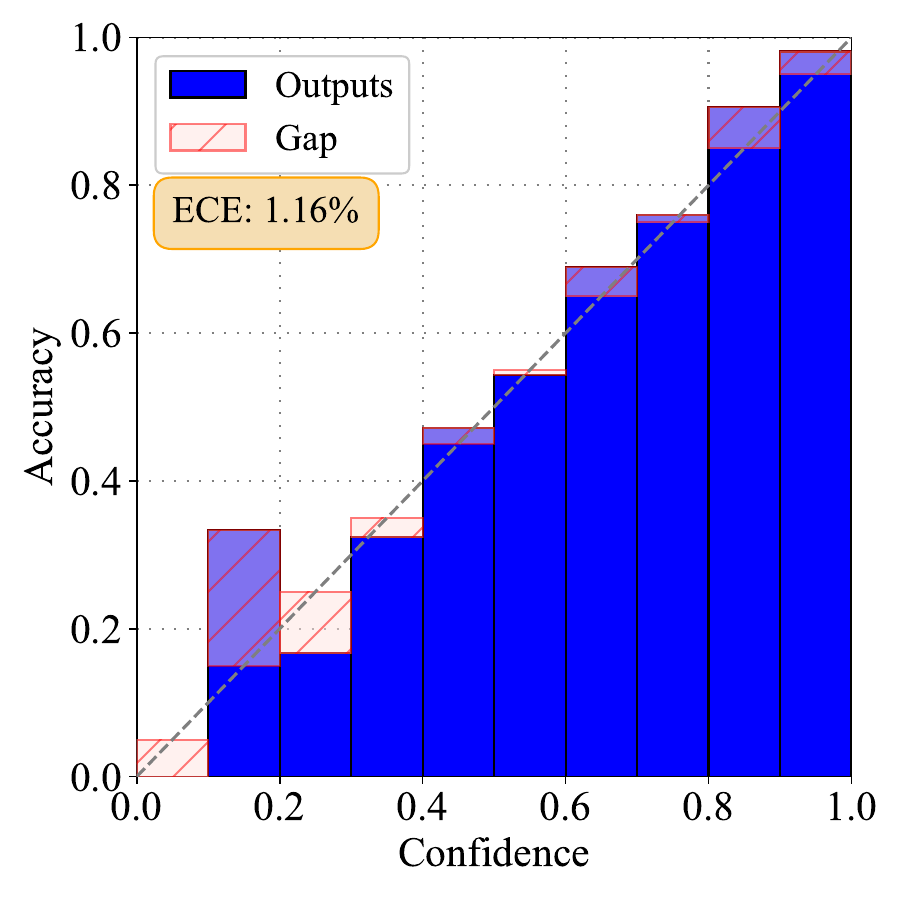} \includegraphics[width=0.49\linewidth]{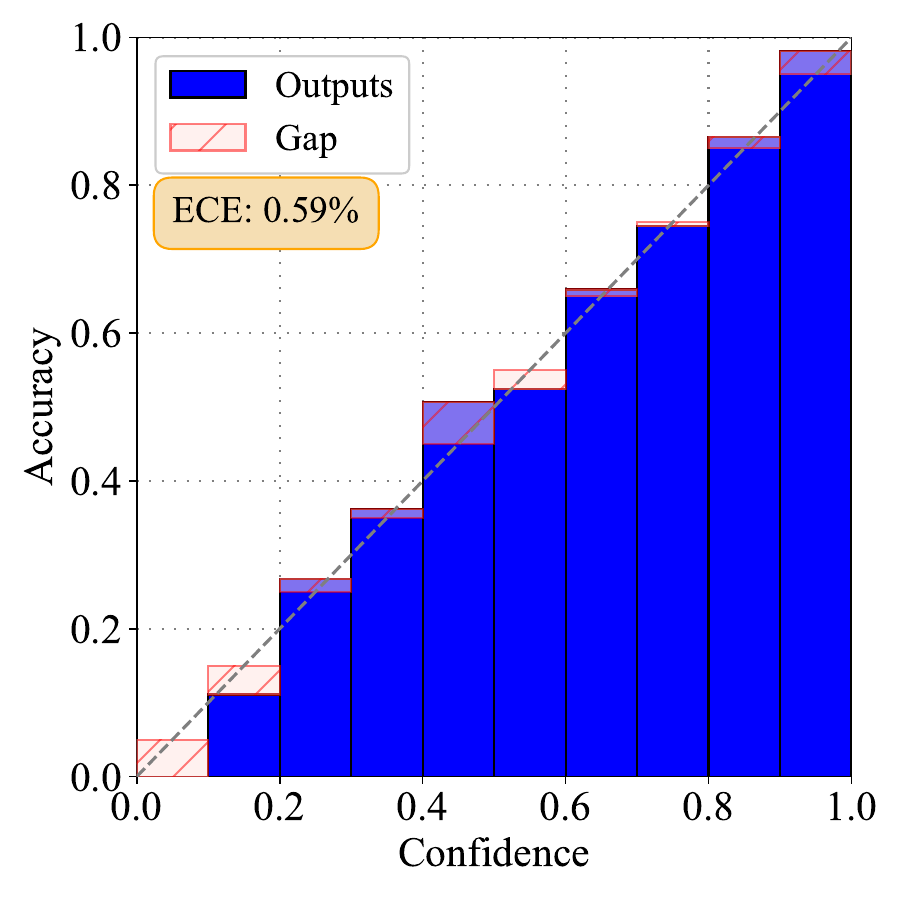}
    \caption{\textcolor{black}{Reliability diagram for FAA-CLIP (\textbf{left} column) vs FedCLIP (\textbf{right} column) using the OfficeHome (\{ACPR\}) dataset. The first row represents the client A, while the second row indicates the global R.}}
    \label{fig:Calibration}
\end{figure}

\begin{figure}
    \centering
    \includegraphics[width=0.49\linewidth]{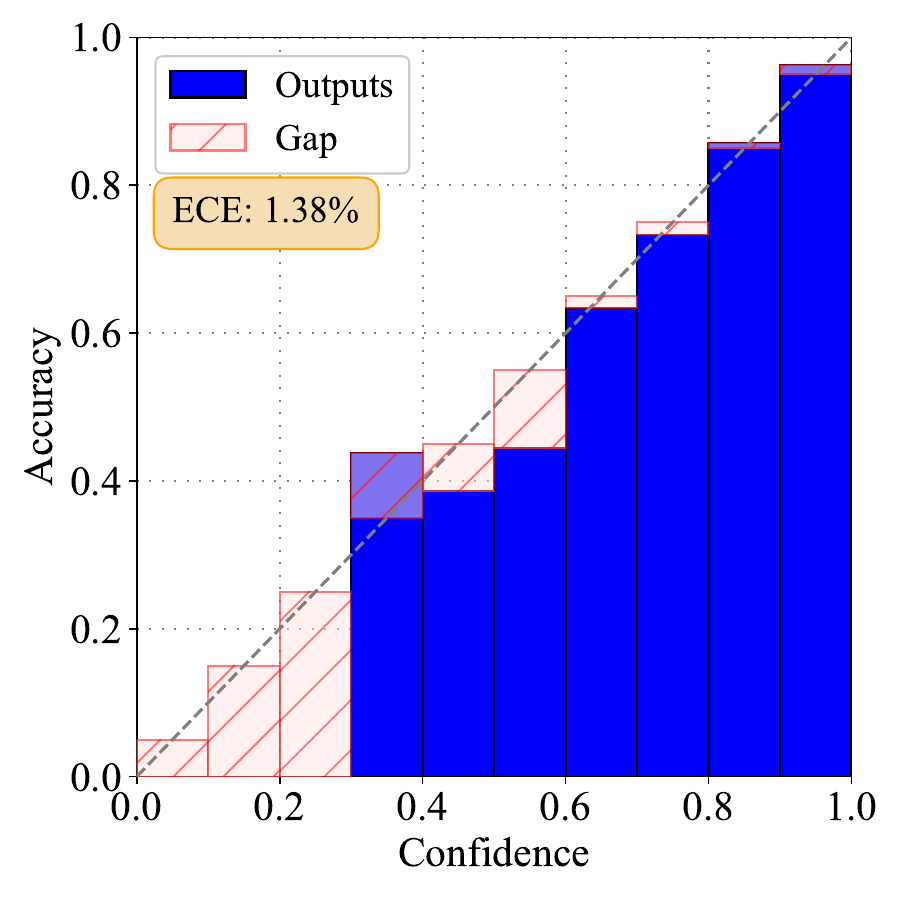} \includegraphics[width=0.49\linewidth]{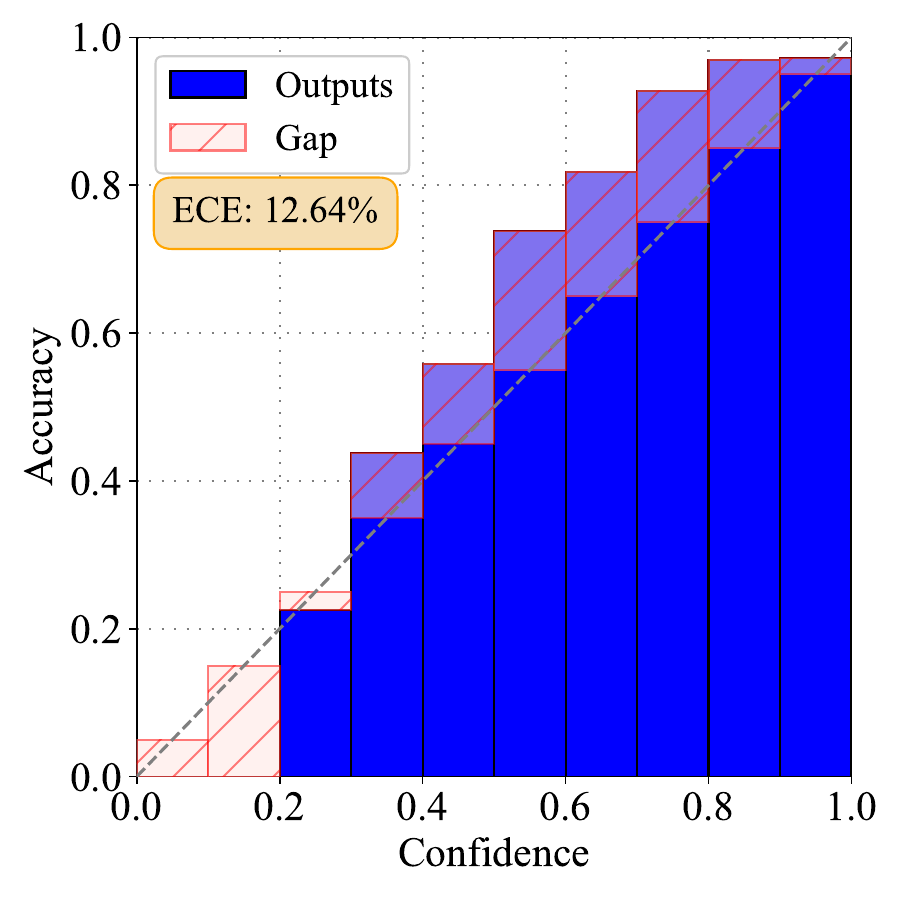}\\
    \includegraphics[width=0.49\linewidth]{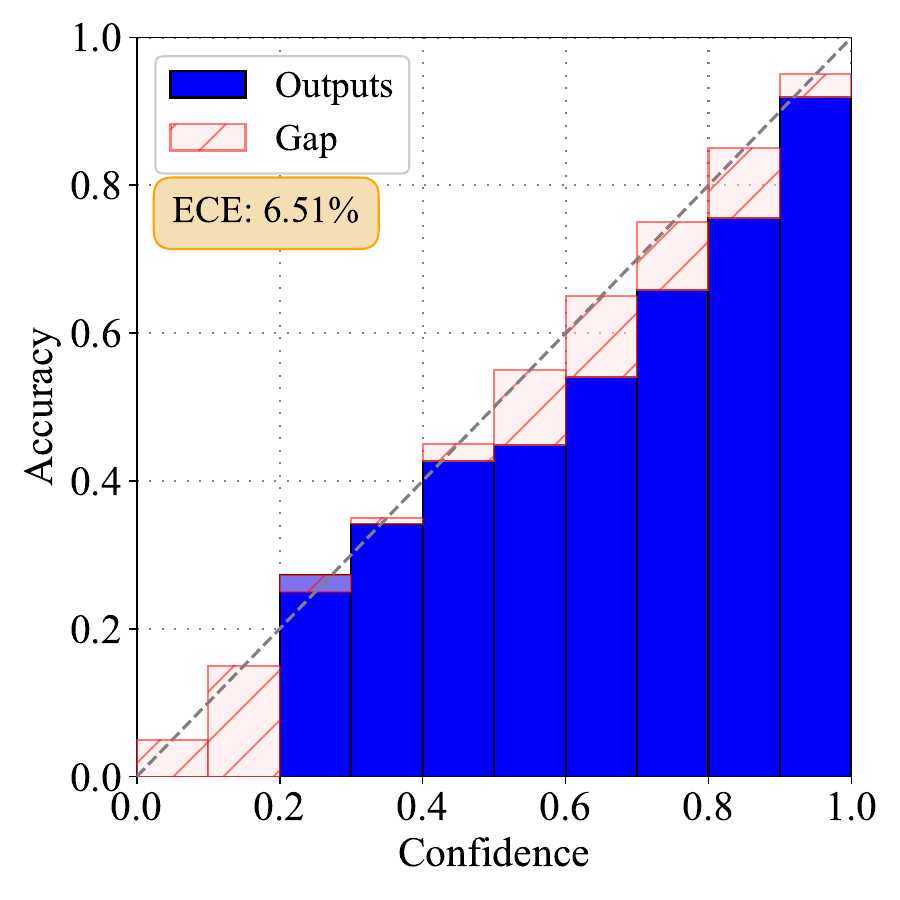} \includegraphics[width=0.49\linewidth]{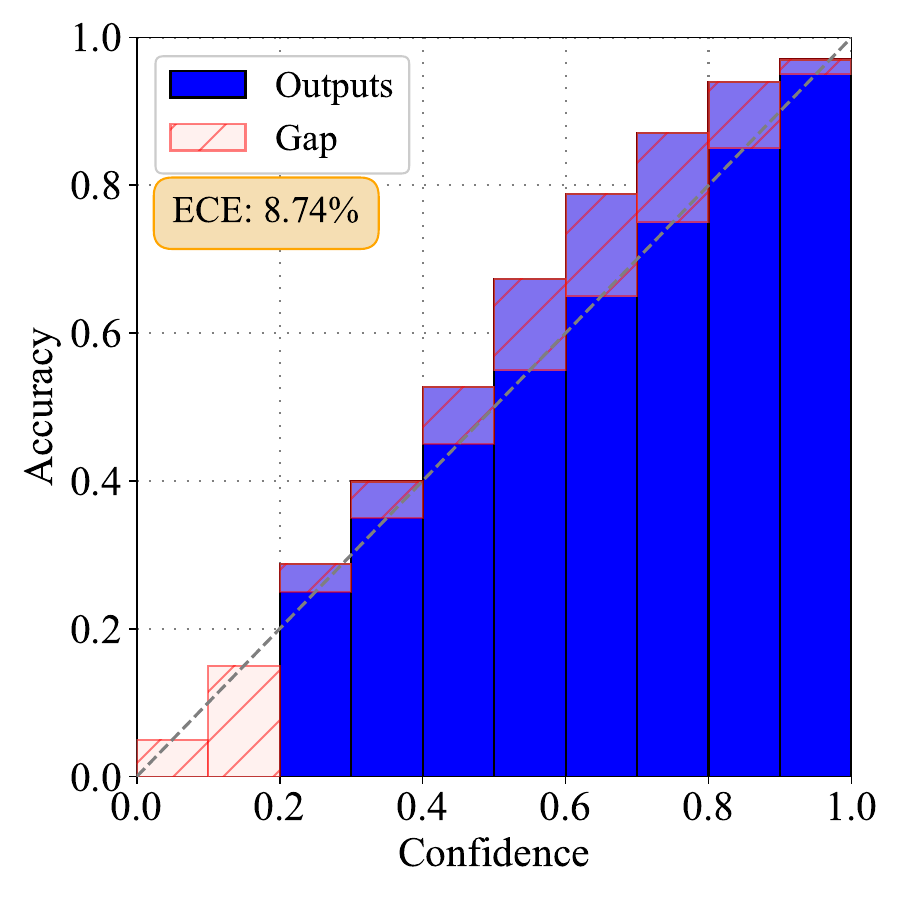}
    \caption{\textcolor{black}{Reliability diagram for FAA-CLIP (\textbf{left} column) vs FedCLIP (\textbf{right} column) on SC dataset. The first row represents the client $C_2$, while the second row indicates the global $C_{glo}$.}}
    \label{fig:Calibration_2}
\end{figure}

\mypar{Calibration analysis.} \textcolor{black}{For each sample of a specific class, the prediction model gives a confidence score that reflects the actual probability of the sample belonging to that class \cite{murugesan2025robust}. For example, calibration of the classification model ensures that the predicted probabilities match the actual probabilities. Following \cite{murugesan2025robust}, the expected calibration error (ECE) metric is used to evaluate the calibration effects. Figure \ref{fig:Calibration} and Figure \ref{fig:Calibration_2} illustrate the reliability plot with the ECE value for FAA-CLIP and FedCLIP using the OfficeHome (\{ACPR\}) and SC datasets. For natural images, FAA-CLIP provides a lower ECE value in the local client (3.12\% vs 4.13\%) while maintaining a similar ECE value in the global compared to FedCLIP (1.16\% vs 0.59\%). However, for medical images, FAA-CLIP reduces the ECE value $\sim 11\%$ on the local client and $\sim 2\%$ on global set compared to FedCLIP, highlighting its potential for the medical domain.}

\begin{figure}[!ht]
\hspace*{-5pt}
    \centering
    \includegraphics[width = 0.485 \textwidth]{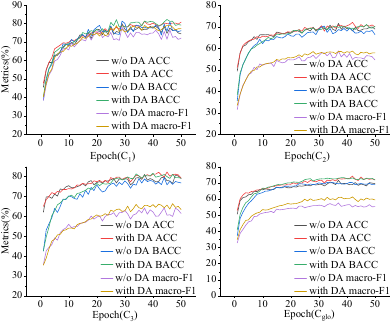}
    \caption{\textcolor{black}{Testing accuracy (ACC), balanced accuracy (BACC) and macro-F1 for each epoch using SC for FAA-CLIP (with DA and without (w/o) DA).}}
    \label{c}
\end{figure}

\subsection{Ablation studies}\label{SS:ablation}

We perform several ablation studies to evaluate the different components of our method. Except for the specific modules, backbones, and parameters described in the following, the configuration of our model is as described in Section \ref{SS:ID}.

% \mypar{Fine-tuning the last MLP layer of \clip{}} \textcolor{red}{We further examined the performance differences between only fine-tuning the final MLP layer and aggregating the parameters of the last MLP layer (referred to as Fed$_{mlp}$) and ours. We use SC dataset for testing and report the testing ACC, BACC and macro-F1 in Table \ref{tab:OfficeHome} and \ref{tab:SCBACC_ablation}. }

\begin{figure}
    \centering
    \includegraphics[width=1\linewidth]{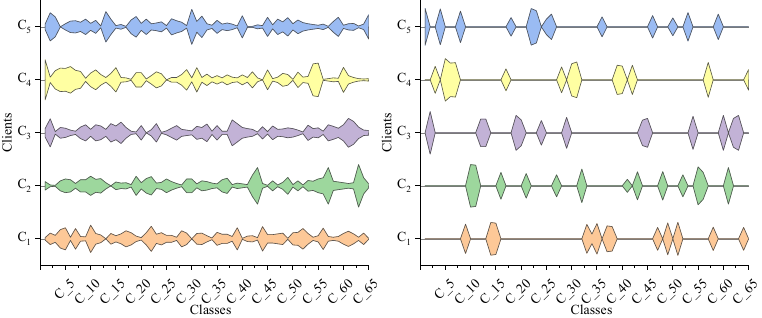}
    \caption{Example of data distribution in each client using kite graph in OfficeHome dataset under practical and pathological setting. $C_1$ to $C_{5}$ indicate clients while $C\_1$ to $C\_65$ represent classes.}
    \label{fig:DataDistribution_OF_practical_pathological}
\end{figure}

\mypar{Impact of the domain adaptation for homogeneous client.} As illustrated in Table \ref{tab:OfficeHome}, the combination of FAM and DA can provide feasible ACC for natural image classification when the client data come from different domains. One concern suggested in \cite{pan2010domain} is that if client data are from the same domain (e.g., the contribution of DA is less), can FAA-\clip{} still exhibit remarkable ACC compared to SOTA methods? Thus, the OfficeHome is used for experiments. For a specific domain (e.g., C), following \cite{qin2023fedapen}, practical and pathological heterogeneity are adopted to allocate data for each client. For practical purposes, data are partitioned into clients based on the Dirichlet distribution. During pathological examination, each client holds $C/N$ (C: Number of classes, N: Number of clients) non-overlapping classes. Then, a different domain (e.g., R) is used for global model evaluation. In our experiments, we use the scikit-learn library to generate Dirichlet distribution randomly to allocate the data, and the number of clients is set to 5 (client domain is R, global domain is C). For pathological, each client holds 13 different classes (derived from domain R), while the global server has three domain data (ACP). Figure \ref{fig:DataDistribution_OF_practical_pathological} shows the data distribution in each client, while Table \ref{tab:OfficeHome(Hemo)} reports the ACC test for the FAA-\clip{} and SOTA methods. As illustrated, the usefulness of DA is limited as FAA-\clip{} indicates a similar ACC testing compared to Fed\clip{} (e.g., 82.69\% AVG vs 82.55\% AVG). However, FAA-\clip{} still outperforms LoRA with large margins (e.g., 82.69\% AVG vs 80.54\% AVG). For pathological, the use of DA demonstrates lower effects for out-of-distribution data (e.g., 73.54\% ACC in domain \{ACP\}), while it can align local data (e.g., 95.16\% ACC in $C_3$).

\mypar{Influence of domain adaptation module.} We validate the usefulness of DA on the SC dataset. Table \ref{tab:MoreAdap} reports the best testing ACC for each client. Furthermore, Figure \ref{c} shows the testing ACC, BACC and F1 of each client for each communication round. As illustrated, model without DA achieves a lower performance compared to the one using DA.  This demonstrates the need for DA to minimize data distribution shifts across clients and global testing set.

\mypar{Impact FAM architecture.} We measured our model's performance when removing different layers within the FAM (the first layer is defined as L1, and so on) to examine the effectiveness of these layers. We used the OfficeHome (Source \{A,C,P\}, Target R), BT and SC datasets for this experiment, and report results in Table \ref{tab:attn}. Removing the batch normalization layer (L2) leads to a performance drop for $C_{glo}$ ACC in the BT and SC datasets. In the OfficeHome dataset, which contains natural images, the impact of batch normalization layers is less significant, and having a larger number of linear layers may result in a performance degradation. It is thus imperative to determine the optimal number of layers.

% Moreover, as Figure \ref{fig:BT_test} SC(b) shows our framework represents a faster convergence in SC data set. 

% \mypar{The impact of data augmentation}
% We examined the commonly used data augmentation methods using the OfficeHome dataset. We first resize the image to $256 \times 256$ then add RandomCrop ($224 \times 224$), followed by Random Horizontally Flip ($p=0.5$). We use the same settings as described in subsection \ref{SS:ID}. Table \ref{tab:OfficeHome} reports the highest test accuracy in the OfficeHome dataset. Using data augmentation, did not show improvements. Even, the use of data augmentation leads to performance degradation.

\mypar{Impact of FAM depth.} \label{SS:deep}
We examine the impact of having a more complex feature adaptation module on performance and efficiency. Thus, we change the FAM to have 8 layers instead of 5: a first linear layer, a batch normalization layer, a LeakyReLU layer, a second linear layer, a batch normalization layer, a LeakyReLU layer, a third linear layer and a softmax activation function. We use the OfficeHome and BT dataset for testing. Table \ref{tab:MoreAdap} reports the highest test ACC for these datasets. As can be seen, the new FAM achieves almost identical average performance compared to the original results. While the performance is higher in domain C, the number of parameters of this deeper FAM is nearly 1.5$\times$ larger than the original one, thereby increasing communication costs.

\mypar{Impact of batch normalization.} Experimental results in \cite{li2021fedbn} have shown the effectiveness of keeping local the batch normalization parameters of each client (not aggregating them in the server) in the context of CNN-based classification. To verify these results, we updated the batch normalization parameters of each client's FAM independently, and used OfficeHome and BT data sets to measure test performance. Table \ref{tab:MoreAdap} gives the highest test ACC for this ablation experiment. As reported, keeping the batch normalization parameters local leads to a worse performance than sharing and aggregating these parameters. This surprising result suggests that local batch normalization is less helpful for FL with VLMs like \clip{}.

\begin{figure}[!ht]
    % \centering
    \includegraphics[width = 1\linewidth]{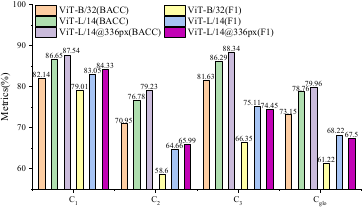}
    % \vspace{-3pt}
    \caption{\textcolor{black}{Testing BACC and macro-F1 of FAA-CLIP for SC dataset, with the ViT-B/32, ViT-L/14 and ViT-L/14@336px backbones.}}
    \label{fig:ab_SC}
\end{figure}

\mypar{Impact of sharing domain classifier parameters.} We studied the influence of aggregating domain classifier $D$ parameters on model performance. Table \ref{tab:MoreAdap} reports the highest test ACC using the OfficeHome and BT datasets. By sharing the parameters of the domain classifier, we gain in performance for domains A and R in OfficeHome. However, for BT, we observe an identical or lower performance compared with the original results.

\mypar{Impact of backbone.} We replaced the ViT-B/32 backbone of the image encoder with pretrained deeper architectures, ViT-L/14 and ViT-L/14@336px, to verify the effectiveness of FAA-\clip{}. We used the SC dataset for these experiments. Figure \ref{fig:ab_SC} shows the best test BACC and macro-F1 for each client and the global server. While it incurs the same communication cost, employing a larger backbones like ViT-L/14 or ViT-L/14@336px leads to a better BACC and F1 score for all clients.

% urs 67.62 65.01 59.08 49.53 68.76 58.67 62.54 52.12 64.5 56.3

% \mypar{Impact of few-shot training.} The performance of VLMs, especially those based on ViTs, usually degrades with fine-tuning is performed with a small amount of data (e.g., overfitting) \cite{gao2024clip}. Thus, we evaluated the generalization ability of our method in a setting where few training samples are given, which often arises in medical imaging applications. Using the SC dataset for this experiment, for each client, we sampled 20\% of the data for local training, and used the remaining 80\% for local testing. Table \ref{tab:SCBACC_ablation} reports the BACC and macro-F1 score for all methods. As can be seen, our FAA-\clip{} largely outperforms other approaches across all clients and performance metrics, demonstrating the robustness of our FAM module design and adaptation techinque.

\begin{table}[]\footnotesize
    \centering
    \renewcommand{\arraystretch}{1}
    \setlength{\tabcolsep}{4.7pt}
    \caption{Testing accuracy (\%) in OfficeHome dataset with non-IID under practical and pathological settings. \textbf{Bold} means the best.}
    \begin{tabular}{cc|ccccc|c|c}
    \toprule
     \multicolumn{2}{c|}{R $\rightarrow$ C}&$C_1$&$C_2$&$C_3$&$C_4$&$C_5$&C&AVG  \\
     \midrule
      \multirow{7}{*}{\rotatebox{90}{Practical}}
      &FedAVG&82.72&87.50&\textbf{87.21}&81.77&80.12&60.11&79.91\\
      &FedProx&79.01&86.31&\textbf{87.21}&85.08&81.33&58.16&79.52 \\
      &MOON&85.80&\textbf{88.10}&84.88&80.66&83.73&60.78&80.66  \\
      &Fed\clip{}&81.48&86.90&83.72&\textbf{92.27}&\textbf{87.95}&62.98&82.55 \\
      &LoRA&\textbf{83.33}&85.71&\textbf{87.21}&86.74&85.54&54.69&80.54  \\
      &PromptFL&38.89&52.38&49.42&48.62&47.59&14.45&41.89 \\
      &\ccol Ours&\ccol 82.10&\ccol 86.90&\ccol 84.30&\ccol 91.71&\ccol \textbf{87.95}&\ccol \textbf{63.21}&\ccol \textbf{82.69}  \\
    \midrule
    \multicolumn{2}{c|}{R $\rightarrow$ ACP}&$C_1$&$C_2$&$C_3$&$C_4$&$C_5$&ACP&AVG  \\
    \midrule
    \multirow{7}{*}{\rotatebox{90}{Pathological}}
    &FedAVG&95.03&96.17&93.55&90.45&89.44&64.28&88.15\\
      &FedProx&\textbf{96.89}&\textbf{96.72}&\textbf{95.70}&96.63&86.96&\textbf{75.32}&91.37 \\
      &MOON&96.27&96.17&93.55&91.01&91.93&66.41&89.22  \\
      &Fed\clip{}&96.27&\textbf{96.72}&94.09&97.19&91.93&74.04&91.71 \\
      &LoRA&94.41&93.44&94.09&93.26&\textbf{93.79}&54.27&87.21  \\
      &PromptFL&85.09&89.62&87.63&85.39&86.34&18.89&75.49\\
      &\ccol Ours&\ccol \textbf{96.89}&\ccol 96.17&\ccol 95.16&\ccol \textbf{97.75}&\ccol \textbf{93.79}&\ccol 73.54&\ccol \textbf{92.22}  \\
    \bottomrule
    \end{tabular}
    
    \label{tab:OfficeHome(Hemo)}
\end{table}

\section{Discussions}
As reported in \cite{gao2024clip}, our results in Table \ref{tab:OfficeHome} and Figure \ref{fig:acc_medical_all} indicate the poor performance of standard VLM fine-tuning in FL. Instead of fine-tuning the whole model, our FAA-\clip{} employs a light-weight adaptation technique based on the proposed FAM. This technique significantly reduces computational needs (e.g., our FAM only needs to transmit nearly $10^5$ parameters) in the case of VLMs like \clip{} \cite{gao2024clip}. The proposed FAM only needs 2.5MB of memory to store the parameters, which is nearly 140$\times$ less than the CLIP encoders with 345MB. Similar to Fed\clip{}, our FAM is a plug-and-play technique that can be easily extended to other vision foundation models or vision language models. As demonstrated in Section \ref{SS:ablation}, the proposed FAM can substantially improve performance in medical imaging datasets, thus bridging the gap with the more common natural image datasets. Furthermore, while sharing FAM paramters can alleviate the problem of domain shifts among clients, using an explicit strategy for DA can further improve the performance of models on unseen domains (e.g., $C_{glo}$ in SC). As stated in Section \ref{SS:deep}, it is also important to maintain a balance between communication costs and parameters. Furthermore, the findings of ROC-AUC and DCA described in Secion \ref{SS:Results} suggest that our method can achieve competitive AUC values with high net benefits, highlighting its potential for clinical applications.

When applied to larger models such as ViT-L/14, our approach achieves a classification F1 score that exceeds 72\% in a large-scale SC classification dataset, thereby providing practical clinical value. However, as reported in our results, sharing the domain classifier $D$ can cause the local model to lose its learned feature representation, thus reducing its performance. This suggests that not all components are suitable for global aggregation; instead, it is recommended to choose the best strategy for aggregation. Experiments on the HK dataset indicate that our method can achieve a comparable performance compared to fine-tuning the entire \clip{} model, with less computational costs. This promotes the development of FL methods for gastrointestinal endoscopy. \textcolor{black}{In addition, FAA-CLIP, while not originally designed for calibration tasks, shows potential for them by achieving a lower ECE value compared to other SOTA methods. This indicates that the FAM itself can be applied to various tasks beyond its initial design.}

% Experiments with limited numbers of samples further demonstrate the opportunity of generalization where data is limited.

% \begin{table}
%     \setlength{\tabcolsep}{5pt}
%         \caption{BACC(\%) and F1 score(\%) in SC. \textbf{Bold} means the best.}
%         \resizebox{\linewidth}{!}{  
%     \begin{tabular}{l|cc|cc|cc|cc|cc}
%     \toprule
%       & \multicolumn{6}{|c|}{Source} & \multicolumn{2}{c}{Target} & 
%      \multicolumn{2}{|c}{\multirow[b]{2}{*}{Avg}}  \\
%      \cmidrule(l{3pt}r{3pt}){2-7} \cmidrule(l{3pt}r{3pt}){8-9}
%    & \multicolumn{2}{c|}{$C_1$} & \multicolumn{2}{c|}{$C_2$ } & \multicolumn{2}{c|}{$C_3$ } & \multicolumn{2}{c|}{$C_{glo}$} & \\
%     \midrule 
%    Fed\clip{} & 50.62 & 47.18 & 47.49 & 36.66 & 52.02 & 40.38 & 49.48 & 39.04 & 49.90 & 40.81 \\
%     FedProx & 51.68 & 49.48 & 46.87 & 40.51 & 52.76 & 46.62 & 47.55 & 42.52 & 49.71 & 44.78  \\
%     FedAVG & 53.05 & 50.10 & 47.49 & 40.69 & 52.89 & 47.74 & 48.24 & 43.45 & 50.41 & 45.49 \\
%     MOON & 51.66 & 47.98 & 47.83 & 40.61 & 52.69 & 46.95 & 48.47 & 42.77 & 50.16 & 44.57  \\
%    \ccol Ours & \ccol \textbf{67.62} & \ccol \textbf{65.01} & \ccol \textbf{59.08} & \ccol \textbf{49.53} & \ccol \textbf{68.76} & \ccol \textbf{58.67} & \ccol \textbf{62.54} & \ccol \textbf{52.12} & \ccol \textbf{64.5} & \ccol \textbf{56.33}  \\
%     \bottomrule
%     \end{tabular}
%     }
%     \label{tab:SCBACC_ablation}
% \end{table}
 
For future work, we intend to integrate FAA-\clip{} into other frameworks and investigate the benefit of having a separate FAM for the text encoder. Exploring more powerful adaptation methods could also improve the performance of our FAA-CLIP model. In what follows, we highlight the strengths and limitations of our approach.
% \textcolor{red}{The proposed plug-and-play FAM can be } 

\mypar{Strengths and limitations.} Our approach is easy to extend (e.g., FAM is easy to customize), and communication efficient compared with traditional FL methods. Furthermore, our method can consistently outperform other FL approaches in commonly used datasets. It also has a faster convergence rate (e.g., Figure \ref{fig:BT_test} shows our method can achieve the same performance with less communication epoch in SC) in medical datasets compared to alternative approaches, demonstrating its effectiveness.

This study used a simple yet effective technique for the feature adaptation. Nevertheless, edge devices like mobile phones may not be able to support the computational expenses related to feature adaptation in other applications such as remote monitoring \cite{longo2023enabling}. We will explore more efficient DA techniques as suggested in \cite{10835760}. Furthermore, for the class imbalance problem that occurs in the medical field, the ACC is not consistent with the F1 score (e.g., $>$10\% performance drop for the F1 score). In the future, we will investigate a contrastive learning strategy specifically designed for rare diseases such as those described in \cite{10330583} to solve this problem. In our experiments, hyper-parameters of FAA-CLIP (e.g., learning rate) were fixed consistently in the local training stage. Using optimizer parameter adjustment (e.g., exponential learning rate regularization) to tune these parameters may help get a better performance.

\section{Conclusion} \label{S:5}
In this paper, we explored the usefulness of VLMs for image classification in the context of FL. To reduce communication costs in FL, we proposed a FAM adapter and only shared the parameters of this module. Furthermore, we introduced a domain adaptation strategy to reduce domain shifts among clients. Experimental results on three natural image datasets and three medical imaging datasets demonstrated the effectiveness of our method compared to state-of-the-art FL approaches. In addition, we conducted a series of ablation studies on various components and design choices, including the FAM architecture and the influence of DA, to better guide researchers in building FL frameworks with VLMs. Our study can help bridge the gap in performance of VLMs in FL scenarios related to medical imaging.

\section*{Acknowledgments}
This research was funded by the National Natural Science Foundation of China \#82260360, the Guilin Innovation Platform and Talent Program \#20222C264164, and the Guangxi Science and Technology Base and Talent Project (\#2022AC18004 and \#2022AC21040).

\bibliographystyle{ieeetr}
\bibliography{ref}

\end{document}